# Taming Numbers and Durations in the Model Checking Integrated Planning System

**Stefan Edelkamp**                                    STEFAN.EDELKAMP@CS.UNI-DORTMUND.DE
*Baroper Straße 301*
*Fachbereich Informatik, GB IV*
*Universität Dortmund*
*44221 Dortmund*
*Germany*

## Abstract

The Model Checking Integrated Planning System (MIPS) is a temporal least commitment heuristic search planner based on a flexible object-oriented workbench architecture. Its design clearly separates explicit and symbolic directed exploration algorithms from the set of on-line and off-line computed estimates and associated data structures.

MIPS has shown distinguished performance in the last two international planning competitions. In the last event the description language was extended from pure propositional planning to include numerical state variables, action durations, and plan quality objective functions. Plans were no longer sequences of actions but time-stamped schedules.

As a participant of the fully automated track of the competition, MIPS has proven to be a general system; in each track and every benchmark domain it efficiently computed plans of remarkable quality. This article introduces and analyzes the most important algorithmic novelties that were necessary to tackle the new layers of expressiveness in the benchmark problems and to achieve a high level of performance.

The extensions include critical path analysis of sequentially generated plans to generate corresponding optimal parallel plans. The linear time algorithm to compute the parallel plan bypasses known NP hardness results for partial ordering by scheduling plans with respect to the set of actions *and* the imposed precedence relations. The efficiency of this algorithm also allows us to improve the exploration guidance: for each encountered planning state the corresponding approximate sequential plan is scheduled.

One major strength of MIPS is its static analysis phase that grounds and simplifies parameterized predicates, functions and operators, that infers knowledge to minimize the state description length, and that detects domain object symmetries. The latter aspect is analyzed in detail.

MIPS has been developed to serve as a complete and optimal state space planner, with admissible estimates, exploration engines and branching cuts. In the competition version, however, certain performance compromises had to be made, including floating point arithmetic, weighted heuristic search exploration according to an inadmissible estimate and parameterized optimization.

## 1. Introduction

Practical action planning and model checking appear to be closely related. The MIPS project targets the integration of model checking techniques into a domain-independent action planner. With the HSF-Spin experimental model checker (Edelkamp, Leue, & Lluch-Lafuente, 2003) we are looking towards the integration of planning technology into an existing model checker. Additional synergies are exploited in the automated compilation of protocol software model checking problems into planner inputs (Edelkamp, 2003).

*Model checking* (Clarke, Grumberg, & Peled, 1999) is the automated process to verify if a formal model of a system satisfies a specified temporal property or not. As an illustrative example, take an elevator control system together with a correctness property that requires an elevator to eventually stop on every call of a passenger or that guarantees that the door is closed, while the elevator is moving. Although the success in checking correctness is limited, model checkers have found many errors in current hardware and software designs. Models often consist of many concurrent sub-systems. Their combination is either synchronous, as often seen in hardware design verification, or asynchronous, as frequently given in communication and security protocols, or in multi-threaded programming languages like Java.

Model checking requires the exploration of very large state spaces containing all reachable system states. This problem is known as the *state explosion problem* and occurs even when the sets of generated states is much smaller than the set of all reachable states.

An error that shows a safety property violation, like a deadlock or a failed assertion, corresponds to one of a set of target nodes in the state space graph. Roughly speaking, *something bad has occured*. A liveness property violation refers to a (seeded) cycle in the graph. Roughly speaking, *something good will never occur*. For the case of the elevator example, eventually reaching a target state where a request button was pressed is a liveness property, while certifying closed doors refers to a safety property.

The two main validation processes in model checking are explicit and symbolic search. In explicit-state model checking each state refers to a fixed memory location and the state space graph is implicitly generated by successive expansions of state.

In symbolic model checking (McMillan, 1993; Clarke, McMillan, Dill, & Hwang, 1992), (fixed-length) binary encodings of system states are used, so that each state can be represented by its characteristic function representation. This function evaluates to true if and only if all Boolean state variables are assigned to bit values with respect to the binary encoding of the system state. Subsequently, the characteristic function is a conjunction of literals with a plain variable for a 1 in the encoding and a negated variable for a 0. Sets of states are expressed as the disjunction of all individual characteristic functions.

The unique symbolic representation of sets of states as Boolean formulae through binary decision diagrams (BDDs) (Bryant, 1992) is often much smaller than the explicit one. BDDs are (ordered) read-once branching programs with nodes corresponding to variables, edges corresponding to variable outcomes, and each path corresponding to an assignment to the variables with the resulting evaluation at the leaves. One reason for the succinctness of BDDs is that directed acyclic graphs may express exponentially many paths. The transition relation is defined on two state variable sets. It evaluates to true, if and only if an operator exists that transforms a state into a valid successor. In some sense, BDDs exploit regularities of the state set and often appear well suited to regular hardware systems. In contrast, many software systems inherit a highly asynchronous and irregular structure, so that the use of BDDs with a fixed variable ordering is generally not flexible enough.

For symbolic exploration, a set of states is combined with the transition relation to compute the set of all possible successor states, i.e. the image. Starting with the initial state, iteration of image computations eventually explores the entire reachable state space.

To improve the efficiency of image computations, transition relations are often provided in partitioned form.

The correspondence between action and model checking (Giunchiglia & Traverso, 1999) can be roughly characterized as follows. Similar to model checkers, action planners implicitly generate large state spaces, and both exploration approaches are based on applying operators to the current state. States spaces in model checking and in planning problems are often modelled as Kripke structures, i.e. state space graphs with states labelled by propositional predicates. The satisfaction of a specified property on the one side corresponds to a complete exploration, and an unsolvable problem on the other side. In this respect, the goal in action planning can be cast as an error with the corresponding trail interpreted as a plan. In the elevator example, the goal of a planning task is to reach a state, in which the doors are open and the elevator is moving. Action planning refers to safety properties only, since goal achievement in traditional and competition planning problems have not yet been extended with temporal properties. However, temporally extended goals are of increasing research interest (Kabanza, Barbeau, & St-Denis, 1997; Pistore & Traverso, 2001; Lago, Pistore, & Traverso, 2002).

In contrast to model checkers that perform either symbolic or explicit exploration, MIPS features both. Moreover, it combines symbolic and explicit search planning in the form of symbolic pattern databases (Edelkamp, 2002b). The planner MIPS implements heuristic search algorithms like A* (Pearl, 1985) and IDA* (Korf, 1985) for exploration, which include state-to-goal approximation into the search process to rank the states to be expanded next. Heuristic search has brought considerable gains to both planning (Bonet & Geffner, 2001; Refanidis & Vlahavas, 2000; Hoffmann & Nebel, 2001; Bertoli, Cimatti, & Roveri, 2001a; Jensen, Bryant, & Veloso, 2002; Feng & Hansen, 2002) and model checking (Yang & Dill, 1998; Edelkamp et al., 2003; Groce & Visser, 2002; Bloem, Ravi, & Somenzi, 2000; Ruys, 2003).

Including resource variables, like the fuel level of a vehicle or the distance between two different locations, as well as action duration are relatively new aspects for competitive planning (Fox & Long, 2003). The input format PDDL2.1 is not restricted to variables of finite domain, but also includes rational (floating-point) variables in both precondition and effects. Similar to a set of atoms described by a propositional predicate, a set of numerical quantities can be described by a set of parameters. Through the notation of PDDL2.1, we refer to parameterized numerical quantities as functions. For example, the fuel level might be parameterized by the vehicle that is present in the problem instance file description.

In the competition, domains were provided in different tracks according to different layers of language expressiveness: *i*) pure propositional planning, *ii*) planning with numerical resources, *iii*) planning with numerical resources and constant action duration, *iv*) planning with numerical resources and variable action duration, and, in some cases, *v*) complex problems usually combining time and numbers in more interesting ways. MIPS competed as a fully automated system and performed remarkably well in all five tracks; it solved a large number of problems and was the only fully automated planner that produced solutions in each track of every benchmark domain.

In this paper the main algorithmic techniques for *taming* rational numbers, objective functions, and action duration are described. The article is structured as follows. First, we review the development of the MIPS system and assert its main contributions. Then we address the object-oriented heuristic search framework of the system. Subsequently, we introduce some terminology that allows us to give a formal definition of the syntax and the semantics of a grounded mixed numerical and propositional planning problem instance. We then introduce the core contributions: critical path scheduling for concurrent plans, and efficient methods for detecting and using symmetry cuts. PERT scheduling produces optimal parallel plans in linear time given a sequence of operators and a precedence relation among them. The paper discusses pruning anomalies and the effect of different optimization criteria. We analyze the correctness and efficiency of symmetry detection in detail. The article closes with related work and concluding remarks.

## 2. The Development of MIPS

The competition version of MIPS refers to initial work (Edelkamp & Reffel, 1999a) in heuristic symbolic exploration of planning domains with the $\mu$cke model checker (Biere, 1997). This approach was effective in sample puzzle solving (Edelkamp & Reffel, 1998) and in hardware verification problems (Reffel & Edelkamp, 1999).

For implementing a propositional planner, we first used our own BDD library called *StaticBdd*, in which large node tables are allocated prior to their use. During the implementation process we changed the BDD representation mainly to improve performance for small planning problems. We selected the public domain `c++` BDD package Buddy (Lind-Nielsen, 1999), which is more flexible. The planning process was semi-automated (Edelkamp & Reffel, 1999b); variable encodings were provided by hand, while the representations of all operators were established by enumerating all possible parameter instances. Once the state space encoding and action transition relation were fixed, exploration in the form of a symbolic breadth-first search of the state-space could be executed. At that time, we were not aware of any other work in BDD-based planning such as the work of Cimatti et al. (1997), which is likely the first link to planning via symbolic model checking. The team used the model checker (nu)SMV as the basis with an atom-to-variable planning state encoding scheme on top of it.

Later on, we developed a parser and a static analyzer to automate the inference of state encodings, the generation of the transition relations, and the extraction of solution paths. In order to minimize the length of the state encoding, the new analyzer clustered atoms into groups (Edelkamp & Helmert, 1999). As confirmed by other attempts (Weismüller, 1998), who started experimenting with PDDL specification in $\mu$cke, state minimization is in fact crucial. The simple encoding using one variable for each atom appears not to be competitive with respect to Graphplan-based (Blum & Furst, 1995) and SAT-plan based planners (Kautz & Selman, 1996). Subsequently, MIPS was the first fully automated planning system based on symbolic model checking technology that could deal with large domain descriptions.

In the second international planning competition MIPS (Edelkamp & Helmert, 2001) could handle the STRIPS (Fikes & Nilsson, 1971) subset of the PDDL language (McDermott, 2000) and some additional features from ADL (Pednault, 1989), namely negative preconditions and (universal) conditional effects. MIPS was one of five planning systems to be awarded for "Distinguished Performance" in the fully automated track. The competition version (Edelkamp & Helmert, 2000) already included explicit heuristic search algorithms based on a bit-vector state representation and the relaxed planning heuristic (RPH) (Hoff-mann & Nebel, 2001) as well as symbolic heuristic search based on the HSP heuristic (Bonet & Geffner, 2001) and a one-to-one atom derivative of RPH. In the competition, we used breadth-first bi-directional symbolic search whenever the single state heuristic search engine got stuck in its exploration.

In between the planning competitions, explicit (Edelkamp, 2001c) and symbolic pattern databases (Edelkamp, 2002b) were proposed as off-line estimators for completely explored problem abstractions. Roughly speaking, pattern database abstractions slice the state vector of fluents into pieces and adjust the operators accordingly. The completely explored subspaces then serve as admissible estimates for the overall search and are competitive with the relaxed planning heuristic in several benchmark domains.

For the third planning competition new levels of the planning domain description language (PDDL) were designed. Level 1 considers pure propositional planning. Level 2 also includes numerical resources and objective functions to be minimized. Level 3 additionally allows the specification of actions with durations. Consequently, MIPS has been extended to cope with these new forms of expressiveness.

First results of MIPS in planning PDDL2.1 problems are presented in (Edelkamp, 2001b). The preliminary treatment illustrates the parsing process in two simple benchmark domains. Moreover, propositional heuristics and manual branching cuts were applied to accelerate sequential plan generation. This work was extended in (Edelkamp, 2002a), where we presented two approximate exploration techniques to bound and to fix numerical domains, first results on symmetry detection based on fact groups, critical path scheduling, an any-time wrapper to produce optimal plans, and a numerical extension to RPH.

## 3. Architecture of MIPS

Figure 1 shows the main components of MIPS and the data flow from the input definition of the domain and the problem instance to the resulting temporal plan in the output. As shown shaded in light gray, MIPS is divided into four parts: pre-compilation, heuristics, search algorithms, and post-compilation (scheduling). Henceforth, the planning process will be coarsely grouped into three stages, pre-compilation, heuristic search planning, and the construction of temporal plans. The problem and domain description files are fed into the system, analyzed and grounded. This fixes the state space problem to be solved. The intermediate result is implicit, but can be saved in a file for use by other planners and model checkers. The basics of pre-compilation are covered in Section 3.2.

The next stage defines the planning process. The object-oriented workbench design of the planner allows different heuristic estimates to be combined with different search strategies and access data structures. Possible choices are listed in Sections 3.3 and 3.4. Temporal planning is based on (PERT) scheduling. This issue of rearranging sequential (relaxed) plans is addressed in detail in Section 4.3.

The planning system was developed in the spirit of the heuristic search framework, HSF for short (Edelkamp, 1999), which allows attachment of newly implemented problem (puzzle) domains to an already compiled system. Similar to the approach that we took in model checking within HSF-Spin, we kept the extensible and general design. In fact we characterized both action planning and protocol validation as single-agent challenges. In contrast to the model checking approach, for planning we devised a hierarchy of system states: the implementation for numerical states is a derived class of the one for propositional states.

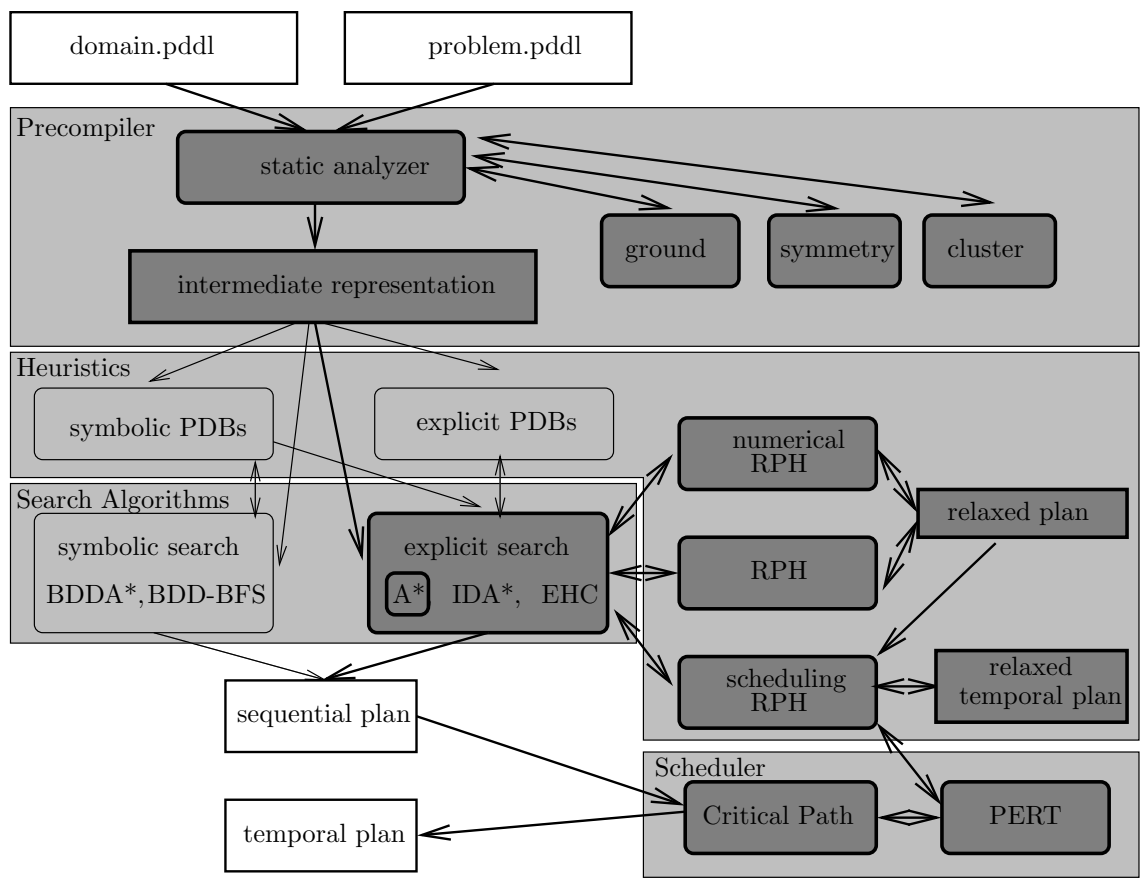

Figure 1: Architecture of MIPS.

Similarly, the heuristic search algorithms are all based on an abstract search class. The main procedures that have to be provided to the search algorithm are a state expansion procedure, and a heuristic search evaluation function, both located in one of the hierarchically organized heuristic estimator classes. In this sense, algorithms in MIPS are general node expanding schemes that can be adapted to very different problems. Additional data structures for the horizon list *Open* and the visited list *Closed* are constructed as parameters of the appropriate search algorithms. As a result, the implementations of the heuristic search algorithms and the associated data structures in the planner MIPS almost match those in our model checker.

## 3.1 Example Problem

The running example for the paper is an instance of a rather simple PDDL2.1 problem in *Zeno-Travel*. It is illustrated in Figure 2. The initial configuration is drawn to the left of the figure and the goal configuration to its right. Some global and local numeric variable assignments are not shown.

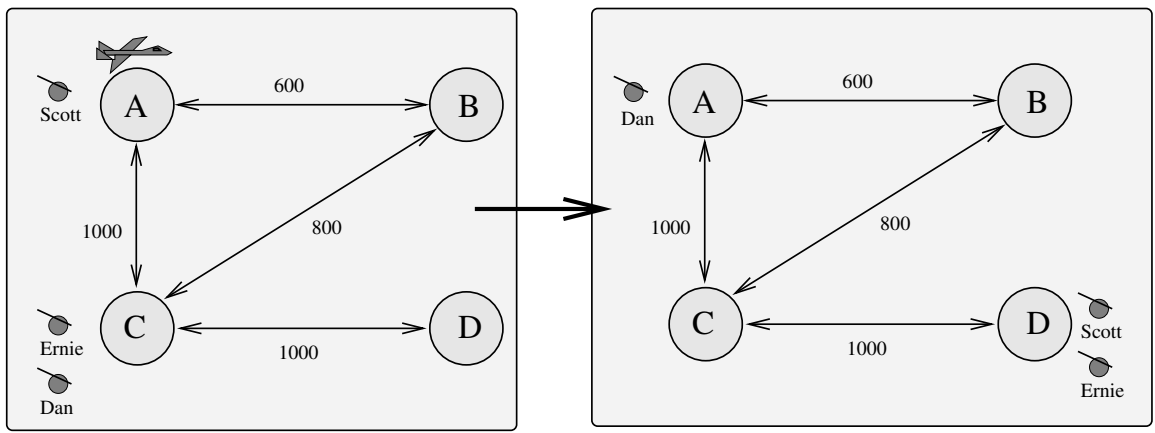

Figure 2: An instance for the *Zeno-Travel* domain with initial state (left) and goal state(s) (right).

Figures 3 and 4 provide the domain and problem specifications[1]. The instance asks for a temporal plan to fly passengers (`dan`, `scott`, and `ernie`) located somewhere on a small map (including the cities `city-a`, `city-b`, `city-c`, and `city-d`) with an aircraft (`plane`) to their respective target destinations. Boarding and debarking take a constant amount of time. The plane has a fixed fuel capacity. Fuel and time are consumed according to the distances between the cities and the travel speed. Fuel can be restored by refueling the aircraft. As a result, the total amount of fuel is also maintained as a numerical quantity.

### 3.2 Precompiler

The static analyzer takes the domain and problem instance as an input, grounds its propositional state information and infers different forms of planner independent static information.

**Parsing** Our simple Lisp parser generates a tree of Lisp entities. It reads the input files and recognizes the domain and problem name. To cope with variable typing, we temporarily assert constant typed predicates to be removed together with other constant predicates in a further pre-compilation step. Thereby, we infer a type hierarchy and an associated mapping of objects to types.

**Indexing** Based on the number of counted objects, indices for the grounded predicates, functions and actions are devised. Since in our example problem we have eight objects and the predicates `at` and `in` have two parameters, we reserve $2 \cdot 8 \cdot 8 = 128$ index positions. Similarly, the function `distance` consumes 64 indices, while `fuel`,

---

1. `[...]` denotes that source fragments were omitted for the sake of brevity. In the given example these are the action definitions for passenger debarking and flying the airplane.

```
(define (domain zeno-travel)
(:requirements :durative-actions :typing :fluents)
(:types aircraft person city)
(:predicates (at ?x - (either person aircraft) ?c - city)
             (in ?p - person ?a - aircraft))
(:functions (fuel ?a - aircraft) (distance ?c1 - city ?c2 - city)
            (slow-speed ?a - aircraft) (fast-speed ?a - aircraft)
            (slow-burn ?a - aircraft)  (fast-burn ?a - aircraft)
            (capacity ?a - aircraft)   (refuel-rate ?a - aircraft)
            (total-fuel-used) (boarding-time) (debarking-time))
(:durative-action board
 :parameters (?p - person ?a - aircraft ?c - city)
 :duration (= ?duration boarding-time)
 :condition (and (at start (at ?p ?c))
                 (over all (at ?a ?c)))
 :effect (and (at start (not (at ?p ?c)))
              (at end (in ?p ?a))))
[...]
(:durative-action zoom
 :parameters (?a - aircraft ?c1 ?c2 - city)
 :duration (= ?duration (/ (distance ?c1 ?c2) (fast-speed ?a)))
 :condition (and (at start (at ?a ?c1))
                 (at start (>= (fuel ?a) (* (distance ?c1 ?c2) (fast-burn ?a)))))
 :effect (and (at start (not (at ?a ?c1)))
              (at end (at ?a ?c2))
              (at end (increase total-fuel-used
                          (* (distance ?c1 ?c2) (fast-burn ?a))))
              (at end (decrease (fuel ?a)
                          (* (distance ?c1 ?c2) (fast-burn ?a))))))
(:durative-action refuel
 :parameters (?a - aircraft ?c - city)
 :duration (= ?duration (/ (- (capacity ?a) (fuel ?a)) (refuel-rate ?a)))
 :condition (and (at start (< (fuel ?a) (capacity ?a)))
                 (over all (at ?a ?c)))
 :effect (at end (assign (fuel ?a) (capacity ?a))))
)
```

Figure 3: Zeno-Travel domain description in PDDL2.1.

slow-speed, fast-speed, slow-burn, fast-burn, capacity, and refuel-rate each reserve eight index positions. For the quantities total-fuel-used, boarding-time, debarking-time only a single fact identifier is needed. Last but not least we model duration as an additional quantity total-time. This special variable is the only one that is overwritten in the least commitment planning approach when scheduling plans as described in Section 4.

```
(define (problem zeno-travel-1)
    (:domain zeno-travel)
    (:objects plane - aircraft
              ernie scott dan - person
              city-a city-b city-c city-d - city)
    (:init (= total-fuel-used 0) (= debarking-time 20) (= boarding-time 30)
           (= (distance city-a city-b) 600)  (= (distance city-b city-a) 600)
           (= (distance city-b city-c) 800)  (= (distance city-c city-b) 800)
           (= (distance city-a city-c) 1000) (= (distance city-c city-a) 1000)
           (= (distance city-c city-d) 1000) (= (distance city-d city-c) 1000)
           (= (fast-speed plane) (/ 600 60)) (= (slow-speed plane) (/ 400 60))
           (= (fuel plane) 750)              (= (capacity plane) 750)
           (= (fast-burn plane) (/ 1 2))     (= (slow-burn plane) (/ 1 3))
           (= (refuel-rate plane) (/ 750 60))
           (at plane city-a) (at scott city-a) (at dan city-c) (at ernie city-c))
    (:goal (and (at dan city-a) (at ernie city-d) (at scott city-d)))
    (:metric minimize total-time)
)
```

Figure 4: Zeno-Travel problem instance.

**Flattening Temporal Identifiers** We interpret each action as an integral entity, so that all timed propositional and numerical preconditions can be merged. Similarly, all effects are merged, independent of time at which they happen. Invariant conditions like (over all (at ?a ?c)) in the action board are added into the precondition set. We discuss the rationale for this step in Section 4.1.

**Grounding Propositions** *Fact-space exploration* is a relaxed enumeration of the planning problem to determine a superset of all reachable facts. Algorithmically, a FIFO fact queue is compiled. Successively extracted facts at the front of the queue are matched to the operators. Each time all preconditions of an operator are fulfilled, the resulting atoms according to the positive effect (add) list are determined and enqueued. This allows us to separate off constant facts from fluents, since only the latter are reached by exploration.

**Clustering Atoms** For a concise encoding of the propositional part we separate fluents into groups, so that each state in the planning space can be expressed as a conjunction of (possibly trivial) facts drawn from each fact group (Edelkamp & Helmert, 1999). More precisely, let $\#p_i(o_1, \ldots, o_{i-1}, o_{i+1}, \ldots, o_n)$ be the number of objects $o_i$ for which the fact $(p\ o_1\ \ldots\ o_n)$ is true. We establish a single-valued invariant at $i$ if $\#p_i(o_1, \ldots, o_{i-1}, o_{i+1}, \ldots, o_n) = 1$. To allow for a better encoding, some predicates like at and in are merged. In the example, three groups determine the unique position of the persons (one of five) and one group determines the position of the plane (one of four). Therefore, $3 \cdot \lceil \log 5 \rceil + 1 \cdot \lceil \log 4 \rceil = 11$ bits suffice to encode the total of 19 fluents.

**Grounding Actions** Fact-space exploration also determines all grounded operators. Once all preconditions are met and grounded, the symbolic effect lists are instantiated. In our case we determine 98 instantiated operators, which, by some further simplifications that eliminate duplicates and trivial operators (no-ops), are reduced to 43.

**Grounding Functions** Simultanous to fact space exploration of the propositional part of the problem, all heads of the numerical formulae in the effect lists are grounded. In the example case only three instantiated formulae are fluent (vary with time): (`fuel plane`) with initial value 750 as well as `total-fuel-used` and `total-time` both initialized with zero. All other numerical predicates are in fact constants that can be substituted in the formula-bodies. In the example, the effect in (`board dan city-a`) reduces to (`increase (total-time) 30`), while (`zoom plane city-a city-b`) has the numerical effects (`increase (total-time) 150`),(`increase (total-fuel-used) 300`)), and (`decrease (fuel plane) 300`). Refuelling, however, does not reduce to a single rational number, for example the effects in (`refuel plane city-a`) only simplify to (`increase (total-time) (/ (- (750 (fuel plane)) / 12.5)))`) and (`assign (fuel plane) 750`). To evaluate the former assignment especially for a forward chaining planner, the variable (`total-time`) has to be instantiated *on-the-fly*. This is due to the fact that the value of the quantity (`fuel plane`) is not constant and itself changes over time.

**Symmetry Detection** Regularities of the planning problem with respect to the transposition of domain objects is partially determined in the static analyzer and is addressed in detail in Section 5.

The intermediate textual format of the static analyzer in annotated grounded PDDL-like representation serves as an interface for other planners or model checkers, and as an additional resource for plan visualization. Figures 5 and 6 show parts of the intermediate representation as inferred in the *Zeno-Travel* example.

### 3.3 Heuristics

MIPS incorporates the following heuristic estimates.

**Relaxed planning heuristic (RPH)** Approximation of the number of planning steps needed to solve the propositional planning problem with all delete effects removed (Hoffmann & Nebel, 2001). The heuristic is constructive, that is it returns the set of operators that appear in the relaxed plan.

**Numerical relaxed planning heuristic (numerical RPH)** Our extension to RPH to deal with with numbers is a combined propositional and numerical approximation scheme allowing multiple operator application.

**Pattern database heuristic (explicit PDB)** Different planning space abstractions are found in a greedy manner, yielding a selection of pattern databases that fit into main memory. In contrast to RPH, pattern database can be designed to be disjoint yielding an admissible estimate as needed for optimal planning in A* (Edelkamp, 2001c).

```
(define (grounded zeno-travel-zeno-travel-1)
  (:fluents
     (at dan city-a)   (at dan city-b)   (at dan city-c)   (at dan city-d)
     (at ernie city-a) (at ernie city-b) (at ernie city-c) (at ernie city-d)
     (at plane city-a) (at plane city-b) (at plane city-c) (at plane city-d)
     (at scott city-a) (at scott city-b) (at scott city-c) (at scott city-d)
     (in dan plane)    (in ernie plane)  (in scott plane))
  (:variables (fuel plane) (total-fuel-used) (total-time))
  (:init
     (at dan city-c)   (at ernie city-c) (at plane city-a) (at scott city-a)
     (= (fuel plane) 750) (= (total-fuel-used) 0) (= (total-time) 0))
  (:goal  (at dan city-a)   (at ernie city-d) (at scott city-d))
  (:metric minimize (total-time) )
  (:group dan
    (at dan city-a)    (at dan city-b)    (at dan city-c)    (at dan city-d)
    (in dan plane))
  (:group ernie
    (at ernie city-a)  (at ernie city-b) (at ernie city-c)  (at ernie city-d)
    (in ernie plane))
  (:group plane
    (at plane city-a)  (at plane city-b) (at plane city-c)  (at plane city-d))
  (:group scott
    (at scott city-a)  (at scott city-b) (at scott city-c)  (at scott city-d)
    (in scott plane))
```

Figure 5: Grounded representation of *Zeno-Travel* domain.

**Symbolic pattern database heuristic (symbolic PDB)** Symbolic PDBs apply to explicit and symbolic heuristic search engines (Edelkamp, 2002b). Due to the succinct BDD-representation of sets of states, symbolic PDBs are often orders of magnitudes larger than explicit ones.

**Scheduling relaxed plan heuristic (scheduling RPH)** Critical-path analysis through scheduling guide the plan finding phase. Like RPH, which computes the length of the greedily extracted sequential plan, scheduling RPH also takes the relaxed sequence of operators into account, but searches for a suitable parallel arrangement, which in turn defines the estimator function.

## 3.4 Exploration Algorithms

The algorithm portfolio includes three main explicit heuristic search algorithms.

**A\*** The A\* algorithm (Hart, Nilsson, & Raphael, 1968) is a variant of Dijkstra's single-source shortest path exploration scheme executed on a re-weighted state space graph. For lower bound heuristics, A\* can be shown to generate optimal plans (Pearl, 1985). Weighting the influence of the heuristic estimate may accelerate solution finding, but also affects optimality (Pohl, 1977).

```
(:action board dan plane city-a
 :condition
    (and (at dan city-a) (at plane city-a))
 :effect
    (and (in dan plane)  (not (at dan city-a))
       (increase (total-time) (30.000000))))
[...]
(:action zoom plane city-a city-b
 :condition
    (and
       (at plane city-a)
       (>= (fuel plane) (300.000000)))
 :effect
    (and (at plane city-b)  (not (at plane city-a))
       (increase (total-time) (60.000000))
       (increase (total-fuel-used) (300.000000))
       (decrease (fuel plane) (300.000000))))
[...]
(:action refuel plane city-a
 :condition
    (and
       (at plane city-a)
       (< (fuel plane) (750.000000)))
 :effect
    (and
       (increase (total-time) (/ (- (750.000000) (fuel plane)) (12.500000)))
       (assign (fuel plane) (750.000000))))
[...]
)
```

Figure 6: Grounded representation of *Zeno-Travel* domain (cont.).

**Iterative-Deepening A\* (IDA\*)** The memory-limited variant of A\* is suited to large exploration problems with evaluation functions of small integer range and low time complexity (Korf, 1985). IDA\* can be extended with bit-state hashing (Edelkamp & Meyer, 2001) to improve duplicate detection with respect to ordinary transposition tables (Reinefeld & Marsland, 1994).

**(Enforced) Hill Climbing (HC)** The approach is another compromise between exploration and exploitation. Enforced HC searches with an improved evaluation in a breadth-first manner and commits established action selections as final (Hoffmann, 2000). Enforced HC is complete in undirected problem graphs.

MIPS also features the following two symbolic search algorithms[2].

---

**Bidirectional Symbolic Breadth-First-Search (BDD-BFS)** The implementation performs bidirectional blind symbolic search, choosing the next search direction to favor the faster execution from the previous iterations (Edelkamp & Helmert, 1999).

**Symbolic A\* (BDDA\*)** The algorithm (Edelkamp & Reffel, 1998) performs guided symbolic search and takes a (possibly partitioned) symbolic representation of the heuristic as an additional input.

### 3.5 Composition of the Competition Version

In Figure 1 we have shaded the parts that were actually used in the competition version of MIPS in dark gray. We used the relaxed planning heuristic for sequential plan generation. The scheduling relaxed planning heuristic was used in temporal domains. Only in Level 2 problems did we use the numerically extended RPH, since it was added to the system in the final weeks of the competition. We experimented with (symbolic) pattern databases with mixed results. Since pattern databases are purely propositional in our implementation and do not provide the retrieval of operators in the optimal abstract plan, we did not include them in the competition version.

Our approach to extend the relaxed planning heuristic with numerical information helps to find plans in challenging numerical domains like *Settlers* and was influenced by Hoffmann's work on his competing planner *Metric-FF* (Hoffmann, 2002a). It builds a relaxed planning graph by computing a fixed-point of a state vector restricted to monotonically increasing propositional and numerical variables. Our version for integrating numbers into the relaxed planning heuristic is not as general as Hoffmann's contribution: it is restricted to variable-to-constant comparisons and lacks the ability to simplify linear constraints. Therefore, we omit the algorithmic details in this paper.

We decided not to employ (enforced) hill climbing for explicit plan generation as is done in Metric-FF. Instead we applied A\* with weight 2, that is the merit for all states $S \in \mathcal{S}$ was fixed as $f(S) = g(S) + 2 \cdot h(S)$. The more conservative plan generation engine was chosen to avoid unrecognized dead-ends, which we expected to be present in benchmark problems. Our objective was that, at least, completeness should be preserved. We also avoided known incomplete pruning rules, like action relevance cuts (Hoffmann & Nebel, 2001) and goal ordering cuts (Koehler & Hoffmann, 2000).

In MIPS, (weighted) A\* accesses both a Dial and a Weak-Heap priority queue data structure. The former is used for propositional planning only, while the latter applies to general planning with scheduling estimates. A Dial priority queue (Dial, 1969) has linear run time behavior, if the maximal value $w(u,v) + h(v) - h(u)$ of all edges $(u,v)$ in the weighted state space graph (labelled with heuristic $h$) is bounded by a constant. Weak-Heaps (Edelkamp & Stiegler, 2002) are simple and efficient relaxations to ordinary heaps. Priority queues have been implemented as dynamic tables that double their sizes if they become filled. Moreover, MIPS stores all generated and expanded states in a hash table with chaining [3]. As a further compression of the planning state space, all variables that appear in the objective function are neglected from hash address calculations and state comparisons. In general, this may lead to a sub-optimal pruning of duplicates. However, for most benchmark domains this will not destroy optimality, since variables addressed in the objective function are frequently monotonic and synonyms found later in the search refer to worse solutions.

---

3. An alternative storage structure is a collection of persistent trees (Bacchus & Kabanza, 2000), one for each predicate. In the best case, queries and update times for the structure are logarithmic in the number of represented atoms.

The price to be paid for selecting A*, especially in planning problems with large branching factors, is that storing all frontier nodes is space consuming. Recent techniques for partial expansion of the horizon list (Yoshizumi, Miura, & Ishida, 2000) or reduced storage of the visited list (Korf & Zhang, 2000; Zhou & Hansen, 2003) have not been included to the system. In most cases, the number of expanded nodes was often not that large, while computing the relaxed planning estimate appeared to be the computational bottleneck. In retrospect, in the domains that were chosen, dead-ends were not central, so that hill climbers appeared to be more effective at finding solutions.

In temporal domains we introduced an additional parameter $\delta$ to scale the influence between propositional estimates ($f_p(S) = g_p(S) + 2 \cdot h_p(S)$) and scheduled ones ($f_s(S) = g_s(S) + 2 \cdot h_s(S)$). More precisely, we altered the comparison function for the priority queue, so that a comparison of parallel length priorities was invoked if the propositional difference of values was not larger than $\delta \in \mathbb{N}_0$. A higher value of $\delta$ refers to a higher influence of the scheduling RPH, while $\delta = 0$ indicates no scheduling at all. In the competition we produced data with $\delta = 0$ (pure MIPS), and $\delta = 2$ (optimized MIPS). In most comparisons of MIPS to other planners the plain version is used, since it produces more solutions.

In (Edelkamp, 2002a) we experimented with an enumeration approach to fix numerical variables to a finite domain, and with an any-time wrapper for optimization of objective functions. These options were excluded from the competition version because of their unpredictable impact on the planner's performance.

## 3.6 Visualization

Visualization is important to ease plan understanding and to quickly detect inefficiencies in the plan generation module. For visualization of plans with MIPS we extended the animation system Vega (Hipke, 2000); a Client-Server architecture that runs an annotated algorithm on the server side, which is visualized on the client side in a Java frontend. The main purpose of the server is to make algorithms accessible through TCP/IP. It is able to receive commands from multiple clients at the same time. We have extended Vega in two ways (cf. Figures 7 and 8).

**Gannt Chart Visualization** Gannt Charts are representations for schedules, in which horizontal bars are drawn for each activity, indicating estimated duration/cost. The user selects any planner to be executed and the domain and problem file, which are interpreted as command line options. Alternatively, established plans can be sent directly to the visualizer with a void planner that merely mirrors the solution file.

**Benchmark Visualization** The second extension is a program suite to visualize all competition domains. At the moment, only sequential plans are shown. For temporal plans, a refined simulation is required, like the one produced by the PDDL2.1 plan validator. Fortunately, in MIPS each temporal plan is a rescheduling of a sequential one.

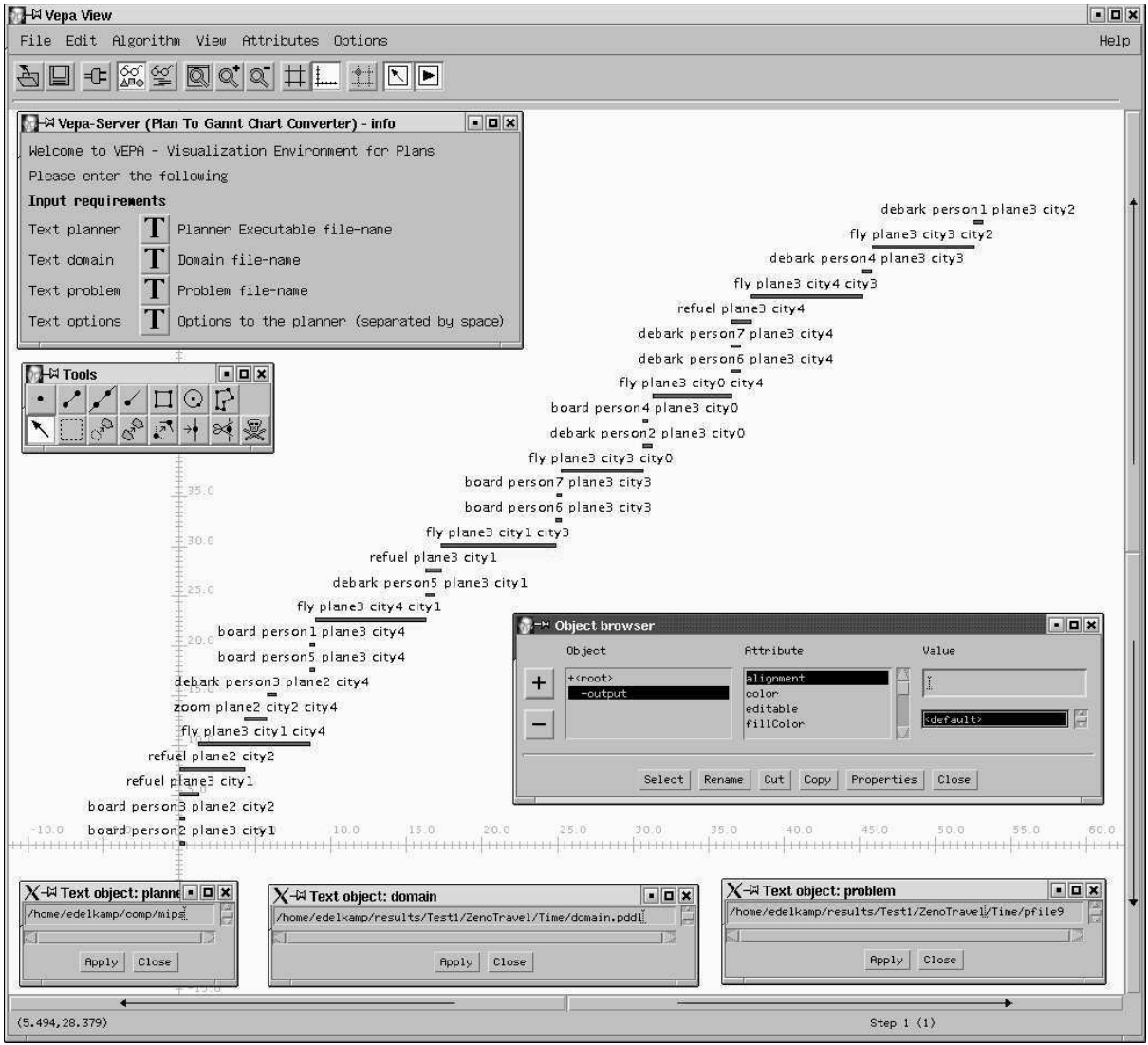

Figure 7: Visualization of a plan in Gannt Chart format.

The images that represent domain objects were collected with an image web search engine[4]. To generalize from specific instances, we advised the MIPS planner to export propositional and numeric state infomation of an established plan in c-like syntax, which in turn is included as a header by the domain visualizer.

## 4. PDDL2.1 Planning

In this section we elaborate on metric and temporal planning in MIPS. We give a formal description on grounded planning instances and introduce the temporal model that we have chosen. Next we look at operator dependency and the resulting action precedence relation. We discuss optimality of the approach and some anomalies that can occur during state space pruning. Last but not least, we turn to the treatment of arbitrary plan objective functions.

---

4. We used *Google* (cf. `www.google.de`) and searched for small GIFs

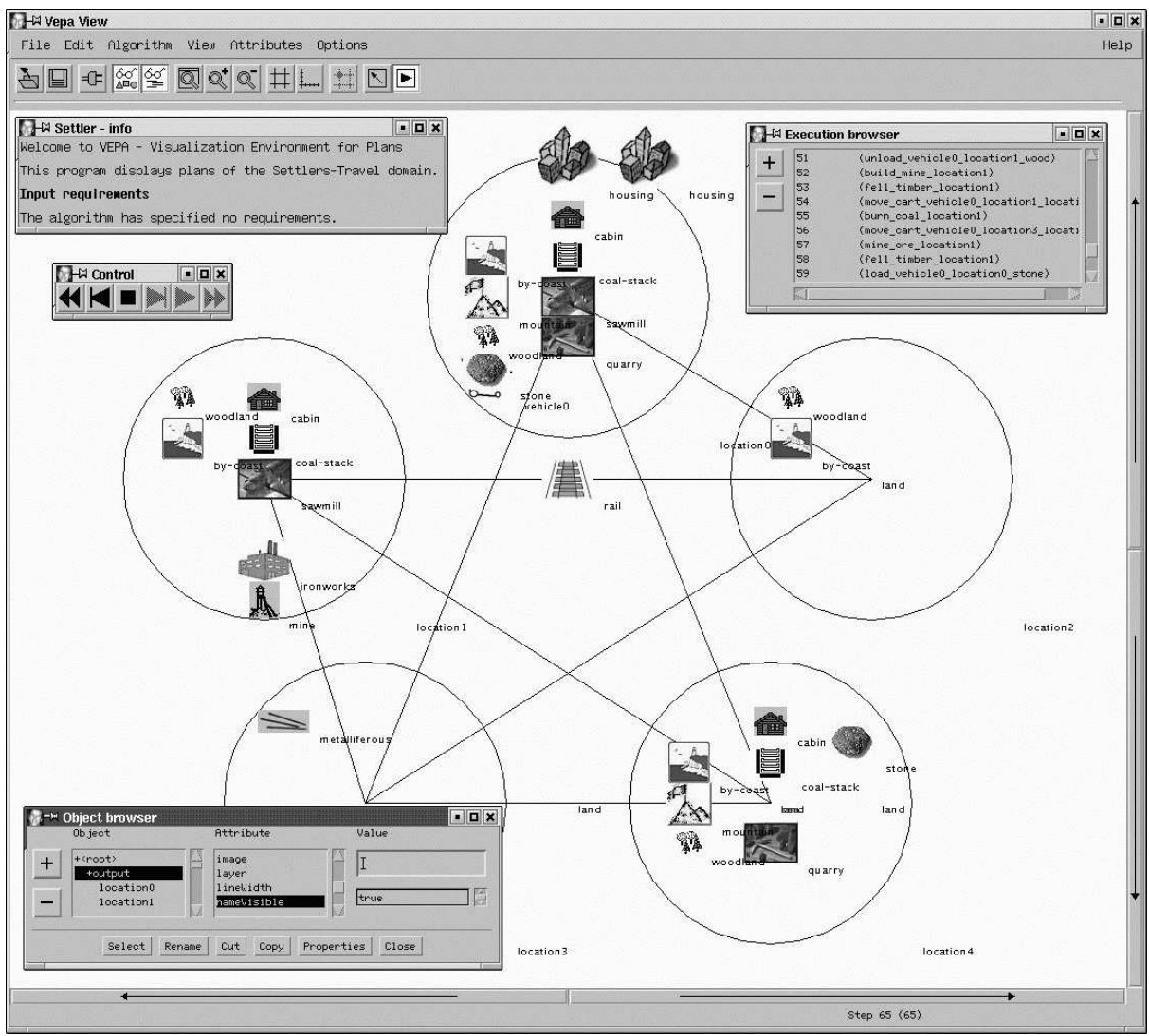

Figure 8: Visualization of a planning problem instance of Settlers.

Table 1 displays the basic terminology for sets used in this paper. As in most currently successful planning system, MIPS grounds parameterized information present in the domain description. For each set we infer a suitable index set, indicated by a bijective mapping $\phi$ from each set to a finite domain. This embedding is important to deal with unique identifiers of entities instead of their textual or internal representation. The arrays containing the corresponding information can then be accessed in constant time.

| Set | Descriptor | Example(s) |
|---|---|---|
| $\mathcal{OBJ}$ | objects | `dan, city-a, plane, ...` |
| $\mathcal{TYPE}$ | object types | `aircraft, person, ...` |
| $\mathcal{PRED}$ | predicates | `(at ?a ?c), (in ?p ?a), ...` |
| $\mathcal{FUNC}$ | functions | `(fuel ?a), (total-time), ...` |
| $\mathcal{ACT}$ | actions | `(board ?a ?p), (refuel ?a), ...` |
| $\mathcal{O}$ | operators | `(board plane scott), ...` |
| $\mathcal{F}$ | fluents/atoms | `(at plane city-b), ...` |
| $\mathcal{V}$ | variables | `(fuel plane), (total-time), ...` |

Table 1: Basic set definitions.

Consequently, like several other planning systems, MIPS refers to grounded planning problem representations.

**Definition 1** *(Grounded Planning Instance) A grounded planning instance is a quadruple $\mathcal{P} = \langle \mathcal{S}, \mathcal{I}, \mathcal{O}, \mathcal{G} \rangle$, where $\mathcal{S}$ is the set of planning states, $\mathcal{I} \in \mathcal{S}$ is the initial state, $\mathcal{G} \subseteq \mathcal{S}$ is the set of goal states. In a mixed propositional and numerical planning problem the state space $\mathcal{S}$ is given by*

$$\mathcal{S} \subseteq 2^{\mathcal{F}} \times I\!\!R^{|\mathcal{V}|},$$

*where $2^{\mathcal{F}}$ is the power set of $\mathcal{F}$. Therefore, a state $S \in \mathcal{S}$ is a pair $(S_p, S_n)$ with propositional part $S_p \in 2^{\mathcal{F}}$ and numerical part $S_n \in I\!\!R^{|\mathcal{V}|}$.*

For the sake of brevity, we assume the operators are in *normal form*, which means that propositional parts (preconditions and effects) satisfy standard STRIPS notation (Fikes & Nilsson, 1971) and numerical parts are given in the form of arithmetic trees $t$ taken from the set of all trees $T$ with arithmetic operations in the nodes and numerical variables and evaluated constants in the leaves. However, there is no fundamental difference for a more general representation of preconditions and effects. The current implementation in MIPS simplifies ADL expressions in the preconditions and takes generic precondition trees for the numerical parts, thereby including comparison symbols, logical operators and arithmetic subtrees[5].

**Definition 2** *(Syntax of Grounded Planning Operators) An operator $o \in \mathcal{O}$ in normal form $o = (\alpha, \beta, \gamma, \delta)$ has propositional preconditions $\alpha \subseteq \mathcal{F}$, propositional effects $\beta = (\beta_a, \beta_d) \subseteq \mathcal{F}^2$ with add list $\beta_a$ and delete list $\beta_d$, numerical preconditions $\gamma$, and numerical effects $\delta$. A numerical precondition $c \in \gamma$ is a triple $c = (h_c, \otimes, t_c)$, where $h_c \in \mathcal{V}$, $\otimes \in \{\leq, <, =, >, \geq\}$, and $t_c \in T$, where $T$ is an arithmetic tree. A numerical effect $m \in \delta$ is a triple $m = (h_m, \oplus, t_m)$, where $h_m \in \mathcal{V}$, $\oplus \in \{\leftarrow, \uparrow, \downarrow\}$ and $t_m \in T$. In this case, we call $h_m$ the head of the numerical effect.*

---

5. In newer versions of MIPS mixing numerical and logical preconditions of the form (`or P (< F 3)`), with P $\in \mathcal{F}$ and F $\in \mathcal{V}$ is in fact feasible. Boolean expressions are put into negational normal form and a disjunction in the precondition will produce different action instantiations.

Obviously, $\otimes \in \{\leq, <, =, >, \geq\}$ represents the associated comparison relation, $\leftarrow$ denotes an assignment to a variable, while $\uparrow$ and $\downarrow$ indicate a respective increase or decrease operation.

**Definition 3** *(Constraint Satisfaction and Modifier Update) Let $\phi$ be the index mapping for variables. A vector $S_n = (S_1, \ldots, S_{|\mathcal{V}|})$ of numerical variables satisfies a numerical constraint $c = (h_c, \otimes, t_c) \in \gamma$ if $S_{\phi(h_c)} \otimes eval(S_n, t_c)$ is true, where $eval(S_n, t_c) \in I\!\!R$ is obtained by substituting all $v \in \mathcal{V}$ in $t_c$ by $S_{\phi(h_c)}$ followed by a simplification of $t_c$.*

*A vector $S_n = (S_1, \ldots, S_{|\mathcal{V}|})$ is* updated *to the vector $S'_n = (S'_1, \ldots, S'_{|\mathcal{V}|})$ by modifier $m = (h_m, \oplus, t_m) \in \delta$, if*

- $S'_{\phi(h_m)} = eval(S_n, t_m)$ *for $\oplus = \leftarrow$,*

- $S'_{\phi(h_m)} = S_{\phi(h_m)} + eval(S_n, t_m)$ *for $\oplus = \uparrow$, and*

- $S'_{\phi(h_m)} = S_{\phi(h_m)} - eval(S_n, t_m)$ *for $\oplus = \downarrow$.*

We next formalize the application of planning operators to a given state.

**Definition 4** *(Semantics of Grounded Planning Operator Application) An operator $o = (\alpha, \beta, \gamma, \delta) \in \mathcal{O}$ applied to a state $S = (S_p, S_n)$, $S_p \in 2^{\mathcal{F}}$ and $S_n \in I\!\!R^{|\mathcal{V}|}$, yields a successor state $S' = (S'_p, S'_n) \in 2^{\mathcal{F}} \times I\!\!R^{|\mathcal{V}|}$ as follows.*

*If $\alpha \subseteq S_p$ and $S_n$ satisfies all $c \in \gamma$ then $S'_p = (S_p \setminus \beta_d) \cup \beta_a$ and the vector $S_n$ is updated for all $m \in \delta$ .*

The propositional update $S'_p = (S_d \setminus \beta_d) \cup \beta_a$ is defined as in standard STRIPS. As an example take the state $S = (S_p, S_n)$ with

$S_p$ = {(at ernie city-d), (at plane city-a), (at scott city-d), (in dan plane)}
$S_n$ = {(fuel plane) : 83.3333, (total-fuel-used) : 1666.6667, (total-time) : 710}.

The successor $S'_n = (S'_p, S'_n)$ of $S$ due to action (debark dan plane city-a) with

$S'_p$ = {(at dan city-a), (at ernie city-d), (at plane city-a), (at scott city-d)}
$S'_n$ = {(fuel plane) : 83.3333, (total-fuel-used) : 1666.6667, (total-time) : 730}.

In some effect lists the order of update operations is important. For example when refuelling the aircraft in *ZenoTravel*, cf. Figure 6, the fuel level has to be reset *after* variable total-time is updated.

The set of goal states $\mathcal{G}$ is often given as $\mathcal{G} = (\mathcal{G}_p, \mathcal{G}_n)$ with a partial propositional state description $\mathcal{G}_p \subset \mathcal{F}$, and $G_n$ as a set of numerical conditions $c = (h_c, \otimes, t_c)$. Moreover, the arithmetic trees $t_c$ usually collapses to simple leaves labelled with numerical constants. Hence, only for the sake of simplifying the complexity analysis for object symmetry we might assume that $|\mathcal{G}_n| \leq |\mathcal{V}|$. Complex goal description are no limitation to the planner, since they can easily transformed to preconditions of an goal-enabling opererator.

## 4.1 Temporal Model

The simplest approach for solving a temporal planning problem is to generate a sequential plan. Of course, this option assumes that the temporal structure contributes only to the value of the plan and not to its correctness. That is, it assumes that there is no necessary concurrency in a valid plan. In cases in which actions achieve conditions at their start points and delete them at their end points, for example, concurrency can be a necessary part of the structure of a valid plan.

**Definition 5** *(Sequential Plan) A solution to a planning problem* $\mathcal{P} = \langle \mathcal{S}, \mathcal{I}, \mathcal{O}, \mathcal{G} \rangle$ *in the form of a* sequential plan $\pi_s$ *is an ordered sequence of operators* $O_i \in \mathcal{O}$, $i \in \{1, \dots, k\}$, *that transforms the initial state* $\mathcal{I}$ *into one of the goal states* $G \in \mathcal{G}$, *i.e., there exists a sequence of states* $S_i \in \mathcal{S}$, $i \in \{0, \dots, k\}$, *with* $S_0 = \mathcal{I}$, $S_k = G$ *such that* $S_i$ *is the outcome of applying* $O_i$ *to* $S_{i-1}$, $i \in \{1, \dots, k\}$.

The time stamp $t_i$ *for a durational operator* $O_i$, $i \in \{1, \dots, k\}$ *is its starting time. If* $d(O_i)$ *is the duration of operator* $O_i$, *then* $t_i = \sum_{j=1}^{i-1} d(O_j)$.

For sequential plans, time stamps are calculated in MIPS using the extra variable `total-time`. This variable is updated when scheduling operators. An example of a sequential plan with time stamps is shown in Figure 12.

Minimizing sequential plan length was the only objective in the first and second planning competitions. Since *Graphplan*-like planners (Blum & Furst, 1995) like IPP (Koehler, Nebel, & Dimopoulos, 1997) and STAN (Long & Fox, 1998) already produced parallel plans (assuming action duration 1), this was indeed a limiting factor in evaluating plan quality. The most important reason for this artificial restriction was that total-ordered plans were easier to automatically validate, a necessity for checking correctness in a competition.

PDDL 2.1 domain descriptions include temporal modifiers *at start*, *over all*, and *at end*, where the label *at start* denotes the preconditions and effects at invocation time of the action, *over all* refers to an invariance condition and *at end* to the finalization conditions and consequences of the action.

In Figure 9 we show two different options for flattening this information to simple preconditions and effects in order to derive the semantic for sequential plans. In the first case (top right), the compound operator is split into three smaller parts, one for action invocation, one for invariance maintenance, and one for action termination. This is the semantics suggested by (Fox & Long, 2003).

In PDDL2.1 there are no effects in the invariance pattern, i.e. $B' = \emptyset$. As in action `board`, it is quite natural to code invariance in the form of conditions ($B$) that perform no actual status change: when a person boards an aircraft in a city the aircraft is required to remain at the city throughout the action. When moving through a corridor, the status of being in the corridor that could be encoded in the invariant would change at the starting time of the action execution.

Moreover, we found that in the benchmarks it is uncommon that new effects in `at-start` are preconditioned for termination control or invariance maintenance, i.e. $A' \cap (B \cup C) = \emptyset$. Even though the intersection of conditions and effects are not formally defined yet, this can be interpreted as executing one construct does not interfere with the other one. This reflects a possible partition of an operator into sub-operators $A$, $B$, $C$, $A'$, $B'$, and $C'$. Dependence and transposition of such separated conditions and effects are considered in Section 4.2.

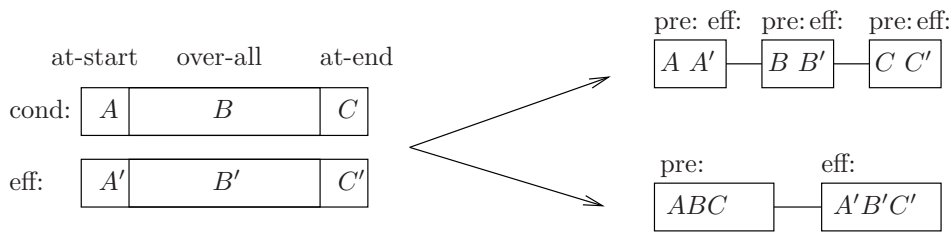

Figure 9: Compiling temporal modifiers into operators.

If we consider the example problem once more, we observe, that in the action `board`, $A'$ consists of the `at` (person airplane) predicate. As seen above, $B$ requires the plane to stay at the city of boarding, while $C$ is empty. In action `zoom`, $A'$ contains the effect that the plane is no longer at the location where the flight started, and $B$ and $C$ are both empty. In all cases we have $A' \cap (B \cup C) = \emptyset$.

If $B' = \emptyset$ and $A' \cap (B \cup C) = \emptyset$ then the sequential execution of the sequence of sub-operators $(A, A', B, B', C, C')$ is equal to the execution sequence $(A, B, C, A', B', C')$. The reasoning is as follows. Since $B' = \emptyset$ we have $(A, A', B, B', C, C') = (A, A', B, C, C')$. Conditions $A' \cap B = \emptyset$ and $A' \cap C = \emptyset$ allows us to exchange the order of the corresponding items, so that $(A, A', B, C, C') = (A, B, C, A', C')$. Once more, we apply $B' = \emptyset$ to derive $(A, B, C, A', C') = (A, B, C, A', B', C')$. The consequence remains valid if the condition $B' = \emptyset$ is weakened to $B' \cap C = \emptyset$.

In MIPS the operator representation at the bottom right of Figure 9 was chosen. Note that the intermediate format of the example problem in Figures 5 and 6 implicitly assumed this temporal model. For sequential planning in the competition benchmark domains we have not observed many deficiencies with this model[6].

However, the applicability of the model for exploiting parallelism is limited. For example consider two people that lift a table from two sides at once, which could not be done with just one person alone. In this case we have a parallel execution of a set of actions that cannot be totally ordered. This is not allowed in MIPS. It may be argued that defining such an action that requires two *different* persons to be at a certain place would require the *equality* construct in PDDL or some form of numerical maintenance of the number of people in the room, but we found another (artificial) example of a planning problem with no total order. Consider the simple STRIPS planning problem domain with $\mathcal{I} = \{B\}$, $\mathcal{G} = \{\{A, C\}\}$, and $\mathcal{O} = \{(\{B\}, \{A\}, \{B\}), (\{B\}, \{C\}, \{B\})\}$. Obviously, both operators are needed for goal achievement, but there is no sequential plan of length 2, since $B$ is deleted in both operators. However, a parallel plan could be devised, since all precondition are fulfilled at the first time step.

---

6. In current versions of MIPS we have refined the model, where *at-start*, *over all*, and *at-end* information is preserved through the grounding process and is attached to each action. The approach does allow dependent operators to overlap and minimizes the number of $\epsilon$ gaps, between *start-start*, *start-end* and *end-end* exclusions. In some of the domains, this improvement yields much better solutions.

## 4.2 Operator Dependency

The definition of operator dependency enables computing optimal schedules of sequential plans with respect to the generated action sequence and its causal operator dependency structure. If all operators are dependent (or void with respect to the optimizer function), the problem is *inherently sequential* and no schedule leads to any improvement.

**Definition 6** *(Dependency/Mutex Relation) Let $L(t)$ denote the set of all leaf variables in the tree $t \in T$. Two grounded operators $O = (\alpha, \beta, \gamma, \delta)$ and $O' = (\alpha', \beta', \gamma', \delta')$ in $\mathcal{O}$ are dependent/mutex, if one of the following three conflicts hold.*

**Propositional conflict** *The propositional precondition set of one operator has a non-empty intersection with the add or the delete list of the other, i.e., $\alpha \cap (\beta'_a \cup \beta'_d) \neq \emptyset$ or $(\beta_a \cup \beta_d) \cap \alpha' \neq \emptyset$.*

**Direct numerical conflict** *The head of a numerical modifier of one operator is contained in some condition of the other one, i.e. there exists a $c' = (h'_c, \otimes, t'_c) \in \gamma'$ and an $m = (h_m, \oplus, t_m) \in \delta$ with $h_m \in L(t'_c) \cup \{h'_c\}$ or there exists a $c = (h_c, \otimes, t_c) \in \gamma$ and an $m' = (h'_m, \oplus, t'_m) \in \delta'$ with $h'_m \in L(t_c) \cup \{h_c\}$.*

**Indirect numerical conflict** *The head of the numerical modifier of one operator is contained in the formula body of the modifier of the other one, i.e., there exists an $m = (h_m, \oplus, t_m) \in \delta$ and $m' = (h'_m, \oplus, t'_m) \in \delta'$ with $h_m \in L(t'_m)$ or $h'_m \in L(t_m)$.*

As an example, the operators (`board scott plan city-a`) and (`fly plane city-a city-c`) have a propositional conflict on the fluent (`at plane city-a`), while (`refuel plane-a city-a`) and (`fly plane city-a city-c`) have a direct numerical conflict on the variable (`fuel plane`). Indirect conflicts are more subtle, and do not appear in the example problem.

We will use dependency to find an optimal concurrent arrangement of the operators in the sequential plan. If $O_2$ is dependent on $O_1$ and $O_1$ appears before $O_2$ in the sequential plan, $O_1$ has to be invoked before $O_2$ starts. The dependence relation is reflexive, i.e. if $O$ is in conflict with $O'$ then $O'$ is in conflict with $O$. Moreover, it appears restrictive when compared to the PDDL 2.1 guidelines for mutual exclusion (Fox & Long, 2003), which allows operators to be partially overlapping even if they are dependent.

However, it is possible to generalize our approach. If, according to the model of Fox and Long, two actions $O_i$ are represented as $(A_i, A'_i, B_i, B'_i, C_i, C'_i)$, $i \in \{1, 2\}$, the dependency violation between $O_1$ and $O_2$ can be located by identifying the sub-operators that interact. In fact we may identify eight possible refined conflicts in which $(A_1 \cup A'_1)$ interacts with $(A_2 \cup A'_2)$, $(A_1 \cup A'_1)$ interacts with $(B_2 \cup B'_2)$, $(A_1 \cup A'_1)$ interacts with $(C_2 \cup C'_2)$, $(B_1 \cup B'_1)$ interacts with $(A_2 \cup A'_2)$, $(B_1 \cup B'_1)$ interacts with $(C_2 \cup C'_2)$, $(C_1 \cup C'_1)$ interacts with $(A_2 \cup A'_2)$, $(C_1 \cup C'_1)$ interacts with $(A_2 \cup A'_2)$, or $(C_1 \cup C'_1)$ interacts with $(A_2 \cup A'_2)$. By asserting duration zero for the pair $(A_i, A'_i)$, $d(A)$ for $(B_i, B'_i)$, and again zero for the pair $(C_i, C'_i)$, one can fix the earliest start and end time of $O_2$ with respect to $O_1$.

In the competition version of MIPS, we stick to the simplified temporal model. For the competition domains, improving sequential plans according to this dependency relation turned out to produce plans of sufficient quality.

In our implementation, the dependence relation is computed beforehand and tabulated for constant time access. To improve the efficiency of pre-computation, the set of leaf variables is maintained in an array, once the grounded operator is constructed.

The original *Graphplan* definition of the propositional mutex relation is close to ours. It fixes interference as $\beta'_d \cap (\beta_a \cup \alpha) \neq \emptyset$ and $(\beta'_a \cup \alpha') \cap \beta_d \neq \emptyset$.

**Lemma 1** *If $\beta_d \subseteq \alpha$ and $\beta'_d \subseteq \alpha'$, operator inference in the* Graphplan *model is implied by the propositional MIPS model of dependence.*

**Proof:** If $\beta_d \subseteq \alpha$ and $\beta'_d \subseteq \alpha'$, for two independent operators $o = (\alpha, \beta)$ and $o' = (\alpha', \beta')$: $\alpha \cap (\beta'_a \cup \beta'_d) = \emptyset$ implies $\beta_d \cap (\beta'_a \cup \beta'_d) = \emptyset$, which in turn yields $\beta_a \cap \beta'_d = \emptyset$. The condition $\beta'_a \cap \beta_d = \emptyset$ is inferred analogously.

The notion of *dependency* is also related to *partial order reduction* in explicit-state model checking (Clarke et al., 1999), where two operators $O_1$ and $O_2$ are *independent* if for each state $S \in \mathcal{S}$ the following two properties hold:

1. Neither $O_1$ or $O_2$ *disable* the execution of the other.

2. $O_1$ and $O_2$ are *commutative*, i.e. $O_1(O_2(S)) = O_2(O_1(S))$ for all $S$.

The next result indicates that both state space enumeration approaches refer to the same property.

**Theorem 1** *(Commutativity) Two independent (STRIPS) operators $O = (\alpha, \beta)$ and $O' = (\alpha', \beta')$ with $\beta_d \subseteq \alpha$ and $\beta'_d \subseteq \alpha'$ are* commutative *and preserve the* enabled *property (i.e. if $O$ and $O'$ are enabled in $S$ then $O$ is enabled in $O'(S)$ and $O'$ is enabled in $O(S)$).*

**Proof:** Since $\beta_d \subseteq \alpha$ and $\beta'_d \subseteq \alpha'$, we have $\beta_a \cap \beta'_d = \emptyset$ and $\beta'_a \cap \beta_d = \emptyset$ by Lemma 1. Let $S'$ be the state $((S \setminus \beta_d) \cup \beta_a)$ and let $S''$ be the state $((S \setminus \beta'_d) \cup \beta'_a)$. Since $(\beta'_a \cup \beta'_d) \cap \alpha = \emptyset$, $O$ is enabled in $S''$, and since $(\beta_a \cup \beta_d) \cap \alpha' = \emptyset$, $O'$ is enabled in $S'$. Moreover,

$$
\begin{aligned}
O(O'(S)) &= (((S \setminus \beta'_d) \cup \beta'_a) \setminus \beta_d) \cup \beta_a \\
&= (((S \setminus \beta'_d) \setminus \beta_d) \cup \beta'_a) \cup \beta_a \\
&= S \setminus (\beta'_d \cup \beta_d) \cup (\beta'_a \cup \beta_a) \\
&= S \setminus (\beta_d \cup \beta'_d) \cup (\beta_a \cup \beta'_a) \\
&= (((S \setminus \beta_d) \setminus \beta'_d) \cup \beta_a) \cup \beta'_a \\
&= (((S \setminus \beta_d) \cup \beta_a) \setminus \beta'_d) \cup \beta'_a = O'(O(S)).
\end{aligned}
$$

As a consequence, operator independence indicates possible transpositions of two operators $O_1$ and $O_2$ to prune exploration in sequential plan generation. A less restrictive notion of independence, in which several actions may occur at the same time even if one deletes an add-effect of another is provided in (Knoblock, 1994). To detect domains for which parallelization leads to no improvement, we utilize the following sufficient criterion.

**Definition 7** (*Inherent Sequential Domains*) *A planning domain is said to be* inherently sequential *if each operator in any sequential plan is either instantaneous (i.e. with zero duration) or dependent on its immediate predecessor.*

The static analyzer checks this by testing each operator pair. While some benchmark domains like *DesertRats* and *Jugs-and-Water* are inherently sequential, others like *ZenoTravel* and *Taxi* are not.

**Definition 8** (*Parallel Plan*) *A solution to a planning problem* $\mathcal{P} = \langle \mathcal{S}, \mathcal{I}, \mathcal{O}, \mathcal{G} \rangle$ *in the form of a* parallel plan $\pi_c = ((O_1, t_1), \ldots, (O_k, t_k))$ *is an arrangement of operators* $O_i \in \mathcal{O}$, $i \in \{1, \ldots, k\}$, *that transforms the initial state* $\mathcal{I}$ *into one of the goal states* $G \in \mathcal{G}$, *where* $O_i$ *is executed at time* $t_i \in I\!R^{\geq 0}$.

An example of a parallel plan for the *ZenoTravel* problem is depicted in Figure 12.

Backstöm (1998) clearly distinguishes partially ordered plans $(O_1, \ldots, O_k, \preceq)$, with the relation $\preceq \, \subseteq \{O_1, \ldots, O_k\}^2$ being a partial order (reflexive, transitive, and antisymmetric), from parallel plans $(O_1, \ldots, O_k, \preceq, \#)$, with $\# \subseteq (\preceq \cup \preceq^{-1})$ (irreflexive, symmetric) expressing, which actions must not be executed in parallel.

**Definition 9** (*Precedence Ordering*) *An ordering* $\preceq_d$ *induced by the operators* $O_1, \ldots, O_k$ *is defined by*

$$O_i \preceq_d O_j : \Longleftrightarrow O_i \text{ and } O_j \text{ are dependent and } 1 \leq i < j \leq k.$$

Precedence is not a partial ordering, since it is neither reflexive nor transitive. By computing the transitive closure of the relation, however, precedence could be extended to a partial ordering. A sequential plan $O_1, \ldots, O_k$ produces an acyclic set of precedence constraints $O_i \preceq_d O_j$, $1 \leq i < j \leq k$, on the set of operators. It is also important to observe, that the constraints are already topologically sorted according to $\preceq_d$ with the index order $1, \ldots, k$.

**Definition 10** (*Respecting Precedence Ordering in Parallel Plan*) *For* $O \in \mathcal{O}$ *let* $d(O) \in I\!R^{\geq 0}$ *be the duration of operator* $O$ *in a sequential plan. In a parallel plan* $\pi_c = ((O_1, t_1), \ldots, (O_k, t_k))$ *that respects* $\preceq_d$, *we have* $t_i + d(O_i) \leq t_j$ *for* $O_i \preceq_d O_j$, $1 \leq i < j \leq k$.

For optimizing plans (Bäckström, 1998) defines *parallel execution time* as $\max\{t_i + d(O_i) \mid O_i \in \{O_1, \ldots, O_k\}\}$, so that if $O_i \preceq O_j$, then $t_i + d(O_i) \leq t_j$, and if $O_i \# O_j$, then either $t_i + d(O_i) \leq t_j$ or $t_j + d(O_j) \leq t_i$. These two possible choices in $\#$ are actually not apparent in our approach, since we already have a precedence relation at hand and just seek the optimal arrangement of operators. Consequently we assert that only one option, namely $t_i + d(O_i) \leq t_j$ can be true, reducing $\#$ to $\preceq_d$. In order to find optimal schedules of sequential plans an approach similar to (Bäckström, 1998) would be necessary. This would dramatically increase the computational complexity, since optimal scheduling of a set of fixed-timed operators is NP-hard. Therefore, we decided to restrict the dependency relation to $\preceq_d$.

**Definition 11** (*Optimal Parallel Plan*) *An optimal parallel plan with respect to a sequence of operators* $O_1, \ldots, O_k$ *and precedence ordering* $\preceq_d$ *is a plan* $\pi^* = ((O_1, t_1), \ldots, (O_k, t_k))$ *with minimal parallel execution time* $OPT = \max\{t_i + d(O_i) \mid O_i \in \{O_1, \ldots, O_k\}\}$ *among all parallel plans* $\pi_c = ((O_1, t'_1), \ldots, (O_k, t'_k))$ *that respect* $\preceq_d$.

**Procedure** *Critical-Path*
**Input:** *Sequence of operators $O_1, \ldots, O_k$, precedence ordering $\preceq_d$*
**Output:** Optimal parallel plan length $\max\{t_i + d(O_i) \mid O_i \in \{O_1, \ldots, O_k\}\}$
    **for all** $i \in \{1, \ldots, k\}$
        $e(O_i) \leftarrow d(O_i)$
        **for all** $j \in \{1, \ldots, i-1\}$
            **if** $(O_j \preceq_d O_i)$
                **if** $e(O_i) < e(O_j) + d(O_i)$
                    $e(O_i) \leftarrow e(O_j) + d(O_i)$
    **return** $\max_{1 \leq i \leq k} e(O_i)$

Figure 10: Algorithm to compute critical path costs.

Many algorithms have been suggested to convert sequential plans into partially ordered ones (Pednault, 1986; Regnier & Fade, 1991; Veloso, Pérez, & Carbonell, 1990). Most of them interpret a totally ordered plan as a maximal constrained partial ordering $\preceq = \{(O_i, O_j) \mid 1 \leq i < j \leq k\}$ and search for less constrained plans. However, the problem of minimum constraint "deordering" has also been proven to be NP-hard, unless the so-called validity check is polynomial (Bäckström, 1998), where deordering maintains validity of the plan by lessening its constrainedness, i.e. $\preceq' \subseteq \preceq$ for a new ordering $\preceq'$.

Since we have an explicit model of dependency and time, optimal parallel plans will not change the ordering relation $\preceq_d$ at all.

### 4.3 Critical Path Analysis

The *Project Evaluation and Review Technique* (PERT) is a critical path analysis algorithm usually applied to project management problems. A critical path is a sequence of activities such that the total time for activities on this path is greater than or equal to any other path of operators. A delay in any tasks on the critical path leads to a delay in the project. The heart of PERT is a network of tasks needed to complete a project, showing the order in which the tasks need to be completed and the dependencies between them.

As shown in Figure 10, PERT scheduling reduces to a variant of Dijkstra's shortest path algorithm in acyclic graphs (Cormen, Leiserson, & Rivest, 1990). As a matter of fact, the algorithm returns the length of the critical path and not the inferred partially ordered plan. However, obtaining the temporal plan is easy. In the algorithm, $e(O_i)$ is the tentative earliest end time of operator $O_i$, $i \in \{1, \ldots, k\}$, while the earliest starting times $t_i$ for all operators in the optimal plan are given by $t_i = e(O_i) - d(O_i)$.

**Theorem 2** *(PERT Scheduling) Given a sequence of operators $O_1, \ldots, O_k$ and a precedence ordering $\preceq_d$, an optimal parallel plan $\pi^* = ((O_1, t_1), \ldots, (O_k, t_k))$ can be computed in optimal time $\mathcal{O}(k + \mid \preceq_d \mid)$.*

**Proof:** The proof is by induction on $i \in \{1, \ldots, k\}$. The induction hypothesis is that after iteration $i$ the value $e(O_i)$ is correct, e.g. $e(O_i)$ is the earliest end time of operator $O_i$. This is clearly true for $i = 1$, since $e(O_1) = d(O_1)$. We now assume that the hypothesis is true $1 \le j < i$ and look at iteration $i$. There are two choices. Either there is a $j \in \{1, \ldots, i-1\}$ with $O_j \preceq_d O_i$. For this case after the inner loop is completed, $e(O_i)$ is set to $\max\{e(O_j) + d(O_j) \mid O_j \preceq_d O_i, j \in \{1, \ldots, i-1\}\}$. On the other hand, $e(O_i)$ is optimal, since $O_i$ cannot start earlier than $\max\{e(O_j) \mid O_j \preceq_d O_i, j \in \{1, \ldots, i-1\}\}$, since all values $e(O_j)$ are already the smallest possible by the induction hypothesis. If there is no $j \in \{1, \ldots, i-1\}$ with $O_j \preceq_d O_i$, then $e(O_i) = d(O_i)$ as in the base case. Therefore, at the end, $\max_{1 \le i \le k} e(O_i)$ is the optimal parallel path length.

The time and space complexity of the algorithm *Critical-Path* are clearly in $\mathcal{O}(k^2)$, where $k$ is the length of the sequential plan. Using an adjacency list representation these efforts can be reduced to time and space proportional to the number of vertices and edges in the dependence graph, which are of size $\mathcal{O}(k + \mid \preceq_d \mid)$. The bound is optimal, since the input consists of $\Theta(k)$ operators and $\Theta(\mid \preceq_d \mid)$ dependencies among them.

Can we apply critical path scheduling, even if we consider the temporal model of Fox and Long, allowing overlapping operator execution of dependent operators? The answer is yes. We have already seen that when considering two dependent operators $O_i$ and $O_j$ in the Fox and Long model, we can determine the earliest start (and end) time of $O_j$ with respect to the fixed start time of $O_i$. This is all that we need. The proof of Theorem 2 shows that we can determine the earliest end time for the operators sequentially.

## 4.4 On the Optimality of MIPS

Since MIPS optimally schedules sequential plans, the question remains, will the system eventually find an optimal plan? In the competition, the system terminates when the first sequential plan is found. Since the relaxed planning heuristic is not admissible, all A\* variants cannot guarantee optimal (sequential or parallel) plans. However, computing optimal plans is desirable, even if – due to limited computational resources – finding optimal plans is hard.

According to our temporal model, in an optimal parallel plan, each operator either starts or ends at the start or end time of another operator. Therefore, at least for a finite number of actions in the optimal plan, we have a possibly exponential but finite number of possible parallel plans.

This immediately leads to the following naive plan enumeration algorithm: For all $|\mathcal{O}|^i$ operator sequences of length $i$, $i \in \mathbb{N}$, generate all possible parallel plans, check for each individual schedule if it transforms the initial state into one of the goals, and take the sequence with smallest parallel plan length. Since all parallel plans are computed, this yields a complete and optimal algorithm. As seen in the example of two persons lifting a table, this approach can be more expressive than applying any algorithm that finds sequential plans first. However, the algorithm is very inefficient.

In practice, the natural assumption is that each parallel plan corresponds to at least one (possible many) sequential one(s). Conversely, each partially ordered plan can be established by generating a totally ordered plan first and then applying a scheduling algorithm to it to find its best partial-order.

The algorithm in the Figure 11 indicates how to wrap a forward chaining planner so that it has *any-time* performance and gradually improves plan quality. The general state

**Procedure** *Any-Time*
**Input:** *Planning Problem* $\langle \mathcal{S}, \mathcal{I}, \mathcal{O}, \mathcal{G} \rangle$
**Output:** Optimal parallel plan length $\alpha$

    $\alpha \leftarrow \infty$
    $Open \leftarrow \mathcal{I}$
    **while** $(Open \neq \emptyset)$
        $S \leftarrow Extract(Open)$
        **for all** $S' \in expand(S)$
            **if** $(S' \in \mathcal{G})$
                $cp \leftarrow Critical\text{-}Path \ (path(S'), \preceq_d)$
                **if** $(cp < \alpha)$
                    $\alpha \leftarrow cp$
            **else**
                $Change(Open, S')$
    **return** $\alpha$

Figure 11: General any-time search algorithm.

expanding scheme maintains the search horizon in the list *Open*. For simplicity the maintenance of stored nodes in the list *Closed* is not shown. In the algorithm, the current best critical path cost $\alpha$ bounds the upcoming exploration process. In turn $\alpha$ is updated each time a plan is found with a shorter critical path.

As in the *CriticalPath* procedure above, the algorithm returns the execution time only, and not the established plan. To compute the plan that meets the returned value $\alpha$, we also store the schedule of the generating sequence $path(S')$ in a global record. In most cases, storing $S'$ is sufficient, since the path and its PERT scheduling can be restored by calling procedure *CriticalPath* at the end of the procedure.

Assuming that each optimal parallel plan is a schedule of a sequential plan and the state space is finite, the any-time extension for a cycle-avoiding enumeration strategy is indeed complete and optimal. The reason for completeness in finite graphs is that the number of acyclic paths in $G$ is finite and with every node expansion, such an algorithm adds new links to its traversal tree. Each newly added link represents a new acyclic path, so that, eventually, the reservoir of paths must be exhausted.

Are there also valid parallel plans that cannot be produced by PERT scheduling of a sequential plan? The answer is no. If a partial ordering algorithm terminates with an optimal schedule, we can generate a corresponding sequential plan while preserving the dependency structure. Optimal PERT-scheduling of this plan with respect to the set of operators and the imposed precedence relation will yield the optimal parallel plan. If all sequential plans are eventually generated, the optimal parallel plan will also be found by PERT scheduling.

The problem of enumeration in infinite state spaces is that there can be infinite plateaus where the plan objective function has a constant value. Normally increasing the length of a plan increases the cost. However, this is not true in all benchmark problems, since there may be an infinite sequence of events that do not contribute to the plan objective. For example, loading and unloading tanks in the pre-competition test domain *DesertRats* does not affect `total-fuel` consumption, which has to be minimized in one of the instances.

Enumeration schemes do not contradict known undecidability results in numerical planning (Helmert, 2002). If we have no additional information like a bound on the maximal number of actions in a plan or on the number of actions that can be executed in parallel, we cannot decide whether a cycle-free enumeration will terminate or not. On the other hand if there is a solution, the any-time algorithm will eventually find it.

### 4.5 Pruning Anomalies

Acceleration techniques like duplicate detection in sequential plan generation have to be chosen carefully to maintain parallel plan length optimality. This approach does affect parallel optimality, as the following example shows. In the *ZenoTravel* problem consider the sequences

```
(zoom city-a city-c plane), (board dan plane city-c),
(refuel plane city-c), (zoom city-c city-a plane),
(board scott plane city-a), (debark dan plane city-a), (refuel plane city-a),
```

and

```
(board scott plane city-a), (zoom city-a city-c plane),
(board dan plane city-c), (refuel plane city-c),
(zoom city-c city-a plane), (debark dan plane city-a), (refuel plane city-a)
```

The two sets of operators are the same and so are the resulting (sequentially generated) states. However, the PERT schedule for the first sequence is shorter than the schedule for the second one, because boarding `scott` can be done in parallel with the final two actions in the plan.

For small problems, such anomalies can be avoided by omitting duplicate pruning. As an example Figure 12 depicts a sequential plan for the example problem instance and its PERT schedule, which turns out to be the overall optimal parallel plan. Another option is to store the resulting parallel plan for state caching instead of the sequential one. Note that in order to ease generation of sequential solutions for large planning problem instances, in the competition version of MIPS we used sequential state pruning.

### 4.6 Heuristic Search

The main drawback of blind path enumeration is that it is seemingly too slow for practical planning. Heuristic search algorithms like A* and IDA* reorder the traversal of states, and (assuming no state caching) do not affect completeness and optimality of the any-time wrapper. The efficiency of the wrapper directly depends on the quality of the path enumeration. In the competition version of MIPS we omitted any-time wrapping, since optimal solutions were not required and the practical run-time behavior is poor.

Instead we used an A* search engine, that terminates on the *first* established solution. The question remains: is there still hope of finding near optimal parallel plans? A general result also applicable for infinite graphs was established by (Pearl, 1985): If the cost of every

```
  0: (zoom plane city-a city-c) [100]          0: (zoom plane city-a city-c) [100]
100: (board dan plane city-c)    [30]        100: (board dan plane city-c)    [30]
130: (board ernie plane city-c)  [30]             (board ernie plane city-c)  [30]
160: (refuel plane city-c)       [40]        100: (refuel plane city-c)       [40]
200: (zoom plane city-c city-a) [100]        140: (zoom plane city-c city-a) [100]
300: (debark dan plane city-a)   [20]        240: (debark dan plane city-a)   [20]
320: (board scott plane city-a)  [30]             (board scott plane city-a)  [30]
350: (refuel plane city-a)       [40]             (refuel plane city-a)       [40]
390: (zoom plane city-a city-c) [100]        280: (zoom plane city-a city-c) [100]
490: (refuel plane city-c)       [40]        380: (refuel plane city-c)       [40]
530: (zoom plane city-c city-d) [100]        420: (zoom plane city-c city-d) [100]
630: (debark ernie plane city-d) [20]        520: (debark ernie plane city-d) [20]
650: (debark scott plane city-d) [20]             (debark scott plane city-d) [20]
```

Figure 12: A sequential plan for *Zeno-Travel* (left) and its PERT schedule (right).

infinite path is unbounded, A*'s cost function $f = g + h$ will preserve optimality. This is additional rationale for choosing an A*-like exploration in MIPS instead of hill climbing or best-first. As in breadth-first search, the rising influence of the $g$-value is crucial.

To find an adequate heuristic estimate for parallel plans is not easy. In fact we have not established a competitive and admissible heuristic, which is required for optimal plan finding in A*. Our choice was a scheduling extension to RPH. In contrast to the RPH, the new heuristic takes the relaxed sequence of operators and searches for a suitable parallel arrangement, which in turn defines the estimator function.

We found that adding PERT-schedules for the path to a state and for the sequence of actions in the relaxed plan is not as accurate as the PERT-schedule of the combined paths. Therefore, the classical merit function of A*-like search engines $f = g + h$ of generating path length $g$ and heuristic estimate $h$ has no immediate correspondence for parallel planning. Consequently, we define the heuristic value of scheduling RPH as the parallel plan length of the combined path minus the parallel plan length of the generating path.

## 4.7 Arbitrary Plan Objectives

In PDDL 2.1 plan metrics other than minimizing total (parallel) execution time can be specified. This influences the inferred solutions. In Figure 13 we depict two plans found by MIPS for the objective functions of minimizing `total-fuel-used`, and minimizing the compound (+ (* 10 (total-time)) (* 1 (total-fuel-used))).

For the first case we computed an optimal value of 1,333.33, while for the second case we established 7,666.67 as the optimized merit. When optimizing time, the ordering of board and zoom actions is important. When optimizing *total-fuel* we reduce speed to save fuel consumption to 333.33 per flight but we may board the first passenger immediately. We also save two refuel actions with respect to the first case.

When increasing the importance of time we can trade refueling actions for time, so that both zooming and flight actions are chosen for the complex minimization criterion.

The first attempt to include arbitrary plan objectives was to alter the PERT scheduling process. However, the results did not match the ones produced by the validator (Long &

```
    0: (board scott plane city-a) [30]
   30: (fly plane city-a city-c) [150]
  180: (board ernie plane city-c) [30]
       (board dan plane city-c)   [30]
  210: (fly plane city-c city-a) [150]
  360: (debark dan plane city-a) [20]
       (refuel plane city-a)    [53.33]
413.33: (fly plane city-a city-c) [150]
563.33: (fly plane city-c city-d) [150]
713.33: (debark ernie plane city-d)[20]
       (debark scott plane city-d)[20]
```

```
    0: (zoom plane city-a city-c) [100]
  100: (board dan plane city-c)   [30]
       (board ernie plane city-c) [30]
       (refuel plane city-c)      [40]
  140: (zoom plane city-c city-a) [100]
  240: (debark dan plane city-a)  [20]
       (board scott plane city-a) [30]
       (refuel plane city-a)      [40]
  280: (fly plane city-a city-c) [150]
  430: (fly plane city-c city-d) [150]
  580: (debark ernie plane city-d) [20]
       (debark scott plane city-d) [20]
```

Figure 13: Optimized plans in *Zeno-Travel* according to different plan objectives.

Fox, 2001), in which the final time is substituted in the objective function after the plan has been built.

The way MIPS evaluates objective functions with time is as follows. First it schedules the (relaxed or final) sequential plan. Variable `total-time` is temporarily substituted for the critical path value and the objective formula is evaluated. To avoid conflicts in subsequent expansions, afterwards value `total-time` is set back to the optimal one in the sequential plan.

## 5. Object Symmetries

An important feature of parameterized predicates, functions and action descriptions in the domain specification file is that actions are transparent to different bindings of parameters to objects. Disambiguating information is only present in the problem instance file.

In the case of typed domains, many planners, including MIPS, compile all type information into additional predicates, attach additional preconditions to actions and enrich the initial states by suitable object-to-type atoms.

As a consequence, a symmetry is viewed as a permutation of objects that are present in the current state, in the goal representation, and transparent to the set of operators.

There are $n!$, $n = |\mathcal{OBJ}|$, possible permutations of the set of objects. Taking into account all type information reduces the number of all possible permutation to

$$\binom{n}{t_1, t_2, \ldots, t_k} = \frac{n!}{t_1! t_2! \ldots t_k!}.$$

where $t_i$ is the number of objects with type $i$, $i \in \{1, \ldots, k\}$. In a moderate sized logistic domain with 10 cities, 10 trucks, 5 airplanes, and 15 packages, this results in $40!/(10! \cdot 10! \cdot 5! \cdot 15!) \geq 10^{20}$ permutations.

To reduce the number of potential symmetries to a tractable size we restrict symmetries to object transpositions, for which we have at most $n(n-1)/2 \in \mathcal{O}(n^2)$ candidates. Using type information this number reduces to

$$\sum_{i=1}^{k} \binom{t_i}{2} = \sum_{i=1}^{k} t_i(t_i - 1)/2.$$

In the following, the set of typed object transpositions is denoted by $\mathcal{SYMM}$. For the Logistics example, we have $|\mathcal{SYMM}| = 45 + 45 + 10 + 105 = 205$.

## 5.1 Generating Object Symmetries for Planning Problems

In this section we compute the subset of $\mathcal{SYMM}$ that includes all object pairs for which the entire planning problem is symmetric. We start with object transpositions for the smallest entities of a planning problem.

**Definition 12** *(Object Transpositions for Fluents, Variables, and Operators) A transposition of objects $(o, o') \in \mathcal{SYMM}$ applied to a fluent $f = (p\ o_1, \ldots, o_{k(p)}) \in \mathcal{F}$, written as $f[o \leftrightarrow o']$, is defined as $(p\ o'_1, \ldots, o'_{k(p)})$, with $o'_i = o_i$ if $o_i \notin \{o, o'\}$, $o_i = o'$ if $o_i = o$, and $o_i = o$ if $o_i = o'$, $i \in \{1, \ldots, k(p)\}$. Object transpositions $[o \leftrightarrow o']$ applied to a variable $v = (f\ o_1, \ldots, o_{k(f)}) \in \mathcal{V}$ or to an operator $O = (a\ o_1, \ldots, o_{k(a)}) \in \mathcal{O}$ are defined analogously.*

For example, in the *ZenoTravel* problem we have (at scott city-a)[scott ↔ dan] = (at dan city-a).

**Lemma 2** *For all $f \in \mathcal{F}$, $v \in \mathcal{V}$, $O \in \mathcal{O}$, and $(o, o') \in \mathcal{SYMM}$: $f[o \leftrightarrow o'] = f[o' \leftrightarrow o]$, $v[o \leftrightarrow o'] = v[o' \leftrightarrow o]$, and $O[o \leftrightarrow o'] = O[o' \leftrightarrow o]$, as well as $f[o \leftrightarrow o'][o \leftrightarrow o'] = f$, $v[o \leftrightarrow o'][o \leftrightarrow o'] = v$, and $O[o \leftrightarrow o'][o \leftrightarrow o'] = O$.*

The brute-force time complexity for computing $f[o \leftrightarrow o'] \in \mathcal{F}$ is of order $\mathcal{O}(k(p))$, where $k(p)$ is the number of object parameters in $p$. However, by pre-computing a $\mathcal{O}(|\mathcal{SYMM}| \cdot |\mathcal{F}|)$ sized lookup table, containing the index of $f' = f[o \leftrightarrow o']$ for all $(o, o') \in \mathcal{SYMM}$, this time complexity can be reduced to $\mathcal{O}(1)$.

**Definition 13** *(Object Transpositions for States) Let $\phi$ be the mapping from set $T$ to $\{1, \ldots, |T|\}$. An object transposition $[o \leftrightarrow o']$ applied to state $S = (S_p, S_n) \in \mathcal{S}$ with $S_n = (v_1, \ldots, v_k)$, $k = |\mathcal{V}|$, written as $S[o \leftrightarrow o']$, is equal to $(S_p[o \leftrightarrow o'], S_n[o \leftrightarrow o'])$ with*

$$S_p[o \leftrightarrow o'] = \{f' \in \mathcal{F} \mid f \in S_p \ \wedge \ f' = f[o \leftrightarrow o']\}$$

*and $S_n[o \leftrightarrow o'] = (v'_1, \ldots, v'_k)$ with $v_i = v'_j$ if $\phi^{-1}(i)[o \leftrightarrow o'] = \phi^{-1}(j)$ for $i, j \in \{1, \ldots, k\}$.*

In the initial state of the example problem we have $\mathcal{I}[\text{dan} \leftrightarrow \text{ernie}] = \mathcal{I}$. The definition for variables is slightly more difficult than for predicates, since, in this case, the variable contents, not just their availability, must match.

The time complexity to compute $S_n[o \leftrightarrow o']$ is $\mathcal{O}(k)$, since testing $\phi^{-1}(i)[o \leftrightarrow o'] = \phi^{-1}(j)$ is available in time $\mathcal{O}(1)$ by building another $\mathcal{O}(|\mathcal{SYMM}| \cdot |\mathcal{V}|)$ sized pre-computed look-up table. Note that these times are worst-case. We can terminate the computation of an object symmetry if a fluent or variable is contradictory. We summarize the complexity results as follows.

**Lemma 3** *The worst-case time complexity to compute $S[o \leftrightarrow o']$ for state $S = (S_p, S_n) \in \mathcal{S}$ and $(o, o') \in \mathcal{SYMM}$ is $\mathcal{O}(|S_p| + |\mathcal{V}|)$ using $\mathcal{O}(|\mathcal{SYMM}| \cdot (|\mathcal{F}| + |\mathcal{V}|))$ space.*

The next step is to lift the concept of object transposition to planning problems.

**Definition 14** *(Object Transpositions for Domains) A planning problem $\mathcal{P} = \langle \mathcal{S}, \mathcal{O}, \mathcal{I}, \mathcal{G} \rangle$ is symmetric with respect to the object transposition $[o \leftrightarrow o']$, abbreviated as $\mathcal{P}[o \leftrightarrow o']$, if $\mathcal{I}[o \leftrightarrow o'] = \mathcal{I}$ and $\forall\ G \in \mathcal{G}$: $G[o \leftrightarrow o'] \in \mathcal{G}$.*

Since goal descriptions are partial, we prefer writing $\mathcal{G}[o \leftrightarrow o'] \in \mathcal{G}$ instead of $\forall\ G \in \mathcal{G}$: $G[o \leftrightarrow o'] \in \mathcal{G}$. Moreover, we assume the goal description complexity for $\mathcal{G}$ to be bounded by $\mathcal{O}(|\mathcal{G}_p| + |\mathcal{V}|)$.

For the *ZenoTravel* problem, the goal descriptor is purely propositional, containing three facts for the target location of `dan`, `ernie`, and `scott`. In the initial state of the running example the planning problem contains no object symmetry, since $\mathcal{I}[\texttt{scott} \leftrightarrow \texttt{ernie}] \neq \mathcal{I}$ and $\mathcal{G}[\texttt{dan} \leftrightarrow \texttt{ernie}] \neq \mathcal{G}$.

Applying Lemma 3 for all $(o, o') \in \mathcal{SYMM}$ yields the time complexity needed to establish all object symmetries.

**Theorem 3** *(Time Complexity for Object Symmetry Detection) The worst-case run-time to determine the set of all object transpositions for which a planning problem $\mathcal{P} = \langle \mathcal{S}, \mathcal{O}, \mathcal{I}, \mathcal{G} \rangle$ is symmetric is $\mathcal{O}(|\mathcal{SYMM}| \cdot (|\mathcal{G}_p| + |\mathcal{I}_p| + |\mathcal{V}|))$.*

## 5.2 Including Goal Symmetry Conditions

Symmetries that are present in the initial state may vanish or reappear during exploration in a forward chaining planner like MIPS. In the *DesertRats* domain, for example, the initial set of supply tanks is indistinguishable so that only one should be loaded into the truck. Once the fuel levels of the supply tanks decrease or tanks are transported to another location, previously existing symmetries are broken. However, when two tanks in one location become empty, they can once again be considered symmetric.

Goal conditions, however, do not change over time, only the initial state $\mathcal{I}$ transforms into the current state $\mathcal{C}$. Therefore, in a pre-compiling phase we refine the set $\mathcal{SYMM}$ to

$$\mathcal{SYMM}' \leftarrow \left\{ (o, o') \in \mathcal{SYMM} \mid \mathcal{G}[o \leftrightarrow o'] = \mathcal{G} \right\}.$$

Usually, $|\mathcal{SYMM}'|$ is much smaller than $|\mathcal{SYMM}|$. For the *ZenoTravel* problem instance, the only object symmetry left in $\mathcal{SYMM}'$ is the transposition of `scott` and `ernie`.

Therefore, we can efficiently compute the set

$$\mathcal{SYMM}''(\mathcal{C}) \leftarrow \left\{ (o, o') \in \mathcal{SYMM}' \mid \mathcal{C}[o \leftrightarrow o'] = \mathcal{C} \right\}$$

of symmetries that are present in the current state. In the initial state $\mathcal{I}$ of the example problem of *Zeno-Travel* we have $\mathcal{SYMM}''(\mathcal{I}) = \emptyset$, but once `scott` and `ernie` share the same location in a state $\mathcal{C}$ this object pair would be included in $\mathcal{SYMM}''(\mathcal{C})$.

The definition requires $\mathcal{C}[o \leftrightarrow o'] = \mathcal{C}$. This does not include symmetric paths from different states. Let $\mathcal{C} = \{(\texttt{at ernie city-c}), (\texttt{at scott city-d})\}$. It is possible that there is a symmetric plan for $\{(\texttt{at ernie city-d}), (\texttt{at scott city-c})\}$ to a common goal. Viewed differently, complex object symmetries of the form $[o_1 \leftrightarrow o_1'][o_2 \leftrightarrow o_2']$ are not detected. For the example we observe $\mathcal{C}[\texttt{scott} \leftrightarrow \texttt{ernie}][\texttt{city-c} \leftrightarrow \texttt{city-d}] = \mathcal{C}$.

With respect to Theorem 3 this additional restriction reduces the time complexity to detect all remaining object symmetries to $\mathcal{O}(|\mathcal{SYMM}'| \cdot (|\mathcal{C}_p| + |\mathcal{V}|))$.

### 5.3 Pruning Operators

If a planning problem with current state $\mathcal{C} \in \mathcal{S}$ is symmetric with respect to the operator transposition $[o \leftrightarrow o']$ then either the application of operator $O \in \mathcal{O}$ or the application of operator $O[o \leftrightarrow o']$ is neglected, significantly reducing the branching factor. Lemma 4 indicates how symmetry is used to reduce exploration.

**Lemma 4** *If operator $O$ is applicable in $S$ and $S = S[o \leftrightarrow o']$ then $O[o \leftrightarrow o']$ is applicable in $S$ and*

$$O(S)[o \leftrightarrow o'] = O[o \leftrightarrow o'](S).$$

**Proof:** If $O$ is applicable in $S$ then $O[o \leftarrow o']$ is applicable in $S[o \leftarrow o']$. Since $S = S[o \leftrightarrow o']$, $O[o \leftrightarrow o']$ is applicable in $S$, and

$$O[o \leftrightarrow o'](S) = O[o \leftrightarrow o'](S[o \leftrightarrow o']) = O(S)[o \leftrightarrow o'].$$

By pre-computing an $\mathcal{O}(|\mathcal{SYMM}| \cdot |\mathcal{O}|)$ sized table the index $\phi(O')$ of operator $O' = O[o \leftrightarrow o']$ can be determined in time $\mathcal{O}(1)$ for each $(o, o') \in \mathcal{SYMM}'$.

**Definition 15** *(Pruning Set) Let $\phi$ be the index mapping from set $T$ to $\{1, \ldots, |T|\}$ and let $\Gamma(\mathcal{C})$ be the set of operators that are applicable in state $\mathcal{C} \in \mathcal{S}$. The pruning set $\Delta(\mathcal{C}) \subset \Gamma(\mathcal{C})$ is defined as the set of all operators that have a symmetric counterpart and that are not of minimal index. The symmetry reduction $\Gamma'(\mathcal{C}) \subseteq \Gamma(\mathcal{C})$ is defined as $\Gamma(\mathcal{C}) \setminus \Delta(\mathcal{C})$.*

**Theorem 4** *(Correctness of Operator Pruning) Reducing the operator set $\Gamma(\mathcal{C})$ to $\Gamma'(\mathcal{C})$ during the exploration of planning problem $\mathcal{P} = \langle \mathcal{S}, \mathcal{O}, \mathcal{I}, \mathcal{G} \rangle$ preserves completeness[7].*

**Proof:** Suppose that for some expanded state $\mathcal{C}$, reducing the operator set $\Gamma(\mathcal{C})$ to $\Gamma'(\mathcal{C})$ during the exploration of planning problem $\mathcal{P} = \langle \mathcal{S}, \mathcal{O}, \mathcal{I}, \mathcal{G} \rangle$ does not preserve completeness. Furthermore, let $\mathcal{C}$ be the state with this property that is maximal in the exploration order.

Then there is a sequential plan $\pi = (O_1 \ldots, O_k)$ in $\mathcal{P}_\mathcal{C} = \langle \mathcal{S}, \mathcal{O}, \mathcal{C}, \mathcal{G} \rangle$ with associated state sequence $(S_0 = \mathcal{C}, \ldots, S_k \subseteq \mathcal{G})$. Obviously, $O_i \in \Gamma(S_{i-1})$, $i \in \{1, \ldots, k\}$. By the choice of $\mathcal{C}$ we have $O_1 \in \Gamma(S_0) \setminus \Gamma'(S_0) = \Delta(S_0)$. By the definition of the pruning set $\Delta(S_0)$ there exists $O_1' = O_1[o \leftrightarrow o']$ of minimal index that is applicable in $S_0$.

Since $\mathcal{P}_\mathcal{C} = \langle \mathcal{S}, \mathcal{O}, \mathcal{C}, \mathcal{G} \rangle = \mathcal{P}_\mathcal{C}[o \leftrightarrow o'] = \langle \mathcal{S}, \mathcal{O}, \mathcal{C}[o \leftrightarrow o'] = \mathcal{C}, \mathcal{G}[o \leftrightarrow o'] = \mathcal{G} \rangle$, we have a sequential plan $O_1[o \leftrightarrow o'], \ldots, O_k[o \leftrightarrow o']$ with state sequence $(S_0[o \leftrightarrow o'] = S_0, S_1[o \leftrightarrow o'], \ldots, S_k[o \leftrightarrow o'] = S_k)$ that reaches the goal $\mathcal{G}$. This contradicts the assumption that reducing the operator set $\Gamma(\mathcal{C})$ to $\Gamma'(\mathcal{C})$ does not preserve completeness for all $\mathcal{C}$.

Since the plan objective refers to instantiated predicates and objects, similar to the initial and goal state, it can be symmetry breaking. In order to preserve optimality, one has to additionally check, to see if the object exchange will influence the plan objective. In practice, objective functions are often based on non-parameterized predicates, in which case an optimal planning algorithm will not be affected by symmetry cuts.

---

7. Generally completeness means that a planner can find any legal plan. This is not what is intended here. We use completeness here in terms of discarding legal plans in favor to equally good symmetric plans.

## 5.4 Symmetry Reduction in MIPS

The main purpose of the restricted implementation in MIPS is to further reduce the run time for object symmetry detection by losing some but not all of its effectiveness. Especially the impact of quantity $\mathcal{O}(|\mathcal{SYMM}'| \cdot |\mathcal{C}_p|)$ for the running time can be considerable.

The key observation is that symmetries are also present in fact groups according to their group representatives. As shown in Figure 5, the fact group of `dan` consists of the facts (`at dan city-a`), (`at dan city-b`), (`at dan city-c`), (`at dan city-d`), and (`in dan plane`). Similarly, `ernie`'s group has facts (`at ernie city-a`), (`at ernie city-a`), (`at ernie city-c`), (`at ernie city-d`), and (`in ernie plane`). The ordering of the facts in the groups can be chosen in a way that, except for the change in the group representative, corresponding facts match. Together with the facts in the groups, the operators that change facts of the groups, are stored in an efficient dictionary.

Therefore, we restrict object transpositions to group representatives. This reduces the set of objects $\mathcal{OBJ}$ that MIPS considers to a considerably smaller subset $\mathcal{OBJ}'$. In the example problem we have $|\mathcal{OBJ}| = 7$, and $|\mathcal{OBJ}'| = 4$. Many objects, e.g. the objects of type `city` in *ZenoTravel*, were not selected as representatives for a single attribute invariance to build a group.

The idea is to obtain a possible transposition of fact group representatives, followed by looking at the respective fact positions of the current and goal state. It may happen, that more than one group has fixed representative $o \in \mathcal{OBJ}'$. In this case, we link groups that have representative $o$ in common. For symmetry detection we test the group chains of both objects for a matching current and goal position.

As above, symmetries based on non-matching goal predicates can be excluded beforehand. Let $\mathcal{RSYMM}$ be the number of remaining symmetries of object representatives. Assume that one representative per group yields a running time per group for propositional object symmetry detection in state $\mathcal{C}$ of $\mathcal{O}(\mathcal{RSYMM} + |\mathcal{C}_p|)$. The remaining comparisons of variables $v \in \mathcal{V}$ are implemented as described in the previous section, but are to be performed only for those object pairs that pass the propositional check.

For pruning operators, MIPS marks all groups that correspond to an object symmetry and that have larger index as *visited*. This guarantees that an operator of at least one group is executed. For each expanded state $S$ and each matching operator $O \in \Gamma(S)$ the algorithm checks, whether an applied operator is present in a visited group, in which case it is pruned. The time complexity is $\mathcal{O}(|\Gamma(S)|)$, since operator group containment can be preprocessed and checked in constant time.

Figure 14 shows the effectiveness of symmetry reduction of the planner MIPS in the *DesertRats* domain, which scales with respect to the total distance $d$, $d \in \{300, 400, 500, 600\}$, that has to be passed ($x$-axis). In the $y$ direction, the number of expanded states in an A* search of MIPS with object symmetry reduction (right bars) and without symmetry reduction (left bars) is shown on a logarithmic scale. As expected, for larger problems symmetry reduction yields performance gains of more than one order magnitude ($d = 500$). It also yields solutions to problems where all algorithms without symmetry reduction fail due to memory restrictions ($d = 600$)[8].

---

8. The memory bound we used for this example was set to 1/2 GByte.

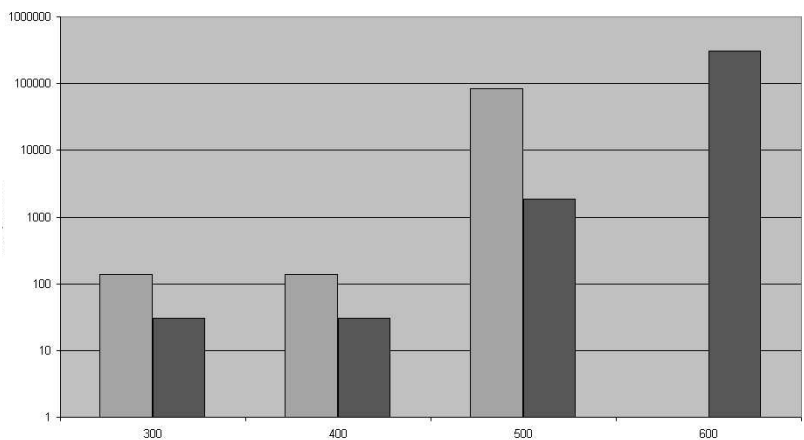

Figure 14: Results in symmetry pruning in *Desert Rats*. Bars show the number of states expanded without/with symmetry detection.

## 6. Related Work

STRIPS problems have been tackled with different planning techniques, most notably by SAT-planning (Kautz & Selman, 1996), IP-planning (Kautz & Walser, 1999), CSP-planning (Rintanen & Jungholt, 1999), graph relaxation (Blum & Furst, 1995), and heuristic search planning (Bonet & Geffner, 2001).

Solving planning problems with numerical preconditions and effects as allowed in Level 2 and Level 3 problems is undecidable in general (Helmert, 2002). However, the structures of the provided benchmark problems are simpler than the general problem class, so these problems are in fact solvable.

### 6.1 Temporal Planning Approaches

The system Metric-FF (Hoffmann, 2002a) extends FF (Hoffmann & Nebel, 2001) as a forward chaining heuristic state space planner for Level 2 problems. Although, MIPS' plan generator shares several ideas with Metric-FF, Hoffmann's system has not yet been extended to deal with temporal domains.

Planner TP4 (Haslum & Geffner, 2001) is in fact a scheduling system based on grounded problem instances. For these cases all formula trees in numerical conditions and assignments reduce to constants. Utilizing admissible heuristics, TP4 minimizes the plan objective of optimal parallel plan length. Our planner has some distinctive advantages: it handles numerical preconditions, instantiates numerical conditions on the fly and can cope with complex objective functions. Besides its input restriction, in the competition, TP4 was somewhat limited by its focus on producing only optimal solutions.

The SAPA system (Do & Kambhampati, 2001) is a domain-independent time and resource planner that can cope with metrics and concurrent actions. SAPA's general expressivity can be judged to be close to that of MIPS. It adapts the forward chaining algorithm of (Bacchus & Ady, 2001). Both planning approaches instantiate actions on the fly and

228

can therefore, in principle, be adapted to handle flexible mixed propositional and numerical planning problems. The search algorithm in SAPA extends partial concurrent plans instead of parallelizing sequential plans. It uses a relaxed temporal planning graph for the yet unplanned events for different heuristic evaluation functions. As an additional feature, SAPA provides the option of specifying deadlines.

The planner LPG (Gerevini & Serina, 2002) is based on local search in planning graphs. It uses a variant of the FF planner for grounding and initial plans are generated through random walk. The subsequent search space of LPG consists of so-called *action graphs* (Gerevini & Serina, 1999). The temporal module performs action graph modifications transforming an action graph into another one. The fast plan generation algorithm in LPG seems to be the best explanation for the speed advantage that LPG has with respect to MIPS, and the higher number of problems LPG solved in some domains. Optimization in LPG is governed by Lagrange multipliers. In temporal domains, actions are ordered using a *precedence graph* that is maintained during search, which uses a more refined dependency relation than ours. This may partly explain why plan quality was in fact consistently better than in MIPS.

IxTeT (Laborie & Ghallab, 1995) is a general constraint-based planning system with its own input format. The planner searches in the space of partial plans and allows general resource and temporal constraints to be posed. The internal representation consists of *chronicles*, with time as a linearly ordered discrete set of instants, and multi-valued state variables that are either rigid or flexible (contingent, controllable, resources), predicates as temporally qualified expressions (events, assertions, resources), and temporal and atemporal constraints. It is not clear how to compare the expressivity of chronicles with PDDL2.1 constructs. This makes it difficult to link the different temporal models and to determine if the technique of critical path scheduling will be applicable to IxTeT or not. In our opinion this is unlikely, since IxTeT is *partial-order*. Note that IxTeT further allows conjunction of predicates, subtasks, constraints and conditional expressions, which are not available in PDDL2.1. The analysis of partial plans that drives the planning process is divided into three different modules: feasibility, satisfiability and resource conflict resolution. In the competition domains IxTeT was not able to compete with local search and heuristic search planners.

HSTS (Muscettola, 1994) is a constraint-based planning system based on temporal activity networks, written in LISP and CRL. At NASA it has been used in many projects like Deep-Space One. It can already represent and reason about metric resources, parallel activities, and general constraints. As in IxTeT the input format is significantly different from PDDL2.1. HSTS has not yet been adapted to represent or reason with conditional branches. However experiences with the HSTS planner showed partial-order planning to be attractive for metric/temporal problems, but with a need for better search control.

Although the PDDL2.1 guidelines in fact do allow infinite branching, the 2002 competition consisted only of finite branching problems. As we indicated earlier, this paper also concentrates on finite branching problems. With finite branching, execution time of an action is fixed, while with infinite branching, a continous range of actions is available.

These problems have been confronted by (real-time) model checking for a long time. Some subclasses of infinite branching problems like timed automata exhibit a finite partitioning through a symbolic representation of states (Pettersson, 1999). By the technique of shortest-path reduction a unique and reduced normal form can be obtained. We have implemented this temporal network structure, since this is the main data structure when exploring timed automata as done by the model checker Uppaal (Pettersson, 1999). For this to work, all constraints must have the form $x_i - x_j \leq c$ or $x_i \leq c$. For example, the set of constraints $x_4 - x_0 \leq -1$, $x_3 - x_1 \leq 2$, $x_0 - x_1 \leq 1$, $x_5 - x_2 \leq -8$, $x_1 - x_2 \leq 2$, $x_4 - x_3 \leq 3$, $x_0 - x_3 \leq -4$, $x_1 - x_4 \leq 7$, $x_2 - x_5 \leq 10$, and $x_1 - x_5 \leq 5$ has the shortest-path reduction $x_4 - x_0 \leq -1$, $x_3 - x_1 \leq 2$, $x_5 - x_2 \leq -8$, $x_0 - x_3 \leq -4$, $x_1 - x_4 \leq 7$, $x_2 - x_5 \leq 10$, and $x_1 - x_5 \leq 5$. If the constraint set is over-constrained, the algorithm will determine unsolvability, otherwise a feasible solution is returned.

Critical path analysis for timed precedence networks is one of the simpler cases for scheduling. We have achieved a simplification by solving the sequential path problem first. Note that many other scheduling techniques apply the presented critical path analysis as a subcomponent (Syslo, Deo, & Kowalik, 1983).

## 6.2 Symmetry Detection in Planning and Model Checkers

Most previous results in symmetry reduction in planning, e.g. (Guéré & Alami, 2001), neglect the combinatorial explosion of possible symmetries or at least assume that the information on existing symmetries in the domain is supplied by the user.

In contrast, our work shares similarities with the approach of Fox & Long (1999,2002) in inferring object symmetry information fully automatically. Fox and Long's work is based on similarities established by the TIM inference module (Fox & Long, 1998). During the search additional information on the current symmetry level in the form of an object transposition matrix is stored and updated together with each state. Our approach is different in the sense that it efficiently computes object symmetries for each state from scratch and it consumes no extra space per node expansion.

Model checking research has a long tradition in symmetry reduction (Clarke et al., 1999). In recent work, Rintanen (2003) connects symmetry detection in planning to model checking approaches for transition systems and SAT solving. Experiments are provided for SAT encodings of the Gripper domain; a prototypical example for symmetry detection. In (Lluch-Lafuente, 2003), our model checker HSF-Spin is extended to effectively combine heuristic search with symmetry detection. It also reflects the fact that (HSF-)Spin's exploration can be modelled using (labelled) transition systems. Positive empirical results are given for non-trivial examples like Peterson's mutual exclusion algorithm and the Database Manager protocol.

We briefly review the fundamental difference between object symmetries (as considered here) and state space symmetries (as considered in model checking).

The latter approach constructs a quotient state space problem ($\mathcal{P}/\sim$) based on a congruence relation, where an equivalence relation $\sim$ of $\mathcal{S}$ is called a *congruence* if for all $s_1, s_2, s_1 \in \mathcal{S}$ with $s_1 \sim s_2$ and operator $O \in \mathcal{O}$ with $O(s_1) = s_1'$ there is an $s_2' \in \mathcal{S}$ with $s_1' \sim s_2'$ and an operator $O' \in \mathcal{O}$ with $O'(s_2) = s_2'$. We have $[O][s] = [s']$ if and only if there is an operator $O \in \mathcal{O}$ mapping $s$ to $s'$ so that $s \in [s]$ and $s' \in [s']$.

A bijection $\phi : \mathcal{S} \to \mathcal{S}$ is said to be a *symmetry* if $\phi(\mathcal{I}) = \mathcal{I}$, $\phi(G) \in \mathcal{G}$ for all $G \in \mathcal{G}$ and for any $s, s' \in \mathcal{S}$ with transition from $s$ to $s'$ there exist a transition from $\phi(s)$ to $\phi(s')$. Any set $A$ of symmetries generates a subgroup $g(A)$ called a symmetry group. The subgroup $g(A)$ induces an equivalence relation $\sim_A$ on states, defined as $s \sim_A s'$ if and only if $\phi(s) = s'$ and $\phi \in g(A)$. Such an equivalence relation is called a *symmetry relation* on $\mathcal{P}$ induced by $A$. The equivalence class of $s$ is called the *orbit* of $s$, denoted as $[s]_A$. Any symmetry relation on $\mathcal{P}$ is a congruence on $\mathcal{P}$. Moreover, $s$ is reachable if and only if $[s]_A$ is reachable from $[\mathcal{I}]_A$. This reduces the search for goal $G \in \mathcal{G}$ to finding state $[G]$.

To explore a state space with respect to a state (space) symmetry, a function *Canonicalize* is needed. Each time a new successor node is generated, it determines a representative element for each equivalence class. Fixing the canonical element is not trivial, so that many systems approximate this normal form. Automatically finding symmetries in this setting is also difficult and can be cast as a computationally hard graph isomorphism problem. Therefore all approaches expect information on the kind of symmetry that is present in the state space graph. One example is a rotational symmetry, defined by a right shift of variables in the state vector.

### 6.3 Model Checking Planners

In the 2000 competition, two other symbolic planners took part: PropPlan (Fourman, 2000), and BDDPlan (Hölldobler & Stör, 2000). Although they did not receive any awards for performance, they show interesting properties. PropPlan performs symbolic forward breadth first search to explore propositional planning problems with propositions for generalized action preconditions and generalized action effects. It performed well in the full ADL Miconic-10 elevator domain (Koehler, 2000). ProbPlan is written in the Poly/ML implementation of SML. BDD-Plan is based on solving the entailment problem in the fluent calculus with BDDs. At that time the authors acknowledged that the concise domain encoding and symbolic heuristic search used in MIPS were improvements.

In the Model-Based Planner MBP the paradigm of planning as symbolic model checking (Giunchiglia & Traverso, 1999) has been implemented for *non-deterministic* planning domains (Cimatti et al., 1998), which can be classified into weak, strong, and strong-cyclic planning, with plans that are represented as state-action tables. For *partially observable* planning, a system is faced with exploring the space of belief states; the power set of the original planning space. Therefore, in contrast to the successor set generation based on action application, observations introduce "And" nodes into the search tree (Bertoli, Cimatti, Roveri, & Traverso, 2001b). Since the approach is a hybrid of symbolic representation of belief states and explicit search within the "And"-"Or" search tree, simple heuristics have been applied to guide the search. The need for heuristics that trade information gain for exploration effort is also apparent in *conformant* planning (Bertoli et al., 2001a). Recent work (Bertoli & Cimatti, 2002) proposes improved heuristics for belief space planning.

The UMOP system parses a non-deterministic agent domain language that explicitly defines a controllable system in an uncontrollable environment (Jensen & Veloso, 2000). The planner also applies BDD refinement techniques such as automated transition function partitioning. New results for the UMOP system extend weak, strong and strong cyclic planning to adversarial planning, in which the environment actively influences the outcome of actions. In fact, the proposed algorithm combines aspects of both symbolic search and game playing. UMOP has not yet participated in a planning competition.

More recent developments in symbolic exploration are expected to influence automated planning in the near future. With SetA*, (Jensen et al., 2002) provide an improved imple-mentation of the symbolic heuristic search algorithm BDDA* (Edelkamp & Reffel, 1998) and Weighted BDDA* (Edelkamp, 2001a). One improvement is that SetA* maintains finer grained sets of states in the search horizon. These are kept in a matrix according to matching $g$- and $h$- values. This contrasts with the plain bucket representation of the priority queue based on $f$-values. The heuristic function is implicitly encoded with value differences of grounded actions. Since sets of states are to be evaluated and some heuristics are state rather than operator dependent it remains to be shown how general this approach is. As above, the planning benchmarks considered are seemingly simple for single-state heuristic search exploration (Hoffmann, 2002b; Helmert, 2001). (Hansen, Zhou, & Feng, 2002) also re-implemented BDDA* and suggest that symbolic search heuristics and exploration algorithms are probably better implemented with algebraic decision diagrams (ADDs). Although the authors achieved no improvement to (Edelkamp & Reffel, 1998) in solving the $(n^2-1)$-Puzzle, the established generalization to guide a symbolic version of the LAO* exploration algorithm (Hansen & Zilberstein, 2001) for *probabilistic* (MDP) planning results in a remarkable improvement in the state-of-the-art (Feng & Hansen, 2002).

## 7. Conclusions

With the competition planning system MIPS, we have contributed a flexible system for a heuristic forward chaining, explicit and symbolic search planner that finds plans in finite-branching numerical problems. The planner parses, pre-compiles, solves, and schedules problem instances, including complex ones with duration, resource variables and different objective functions. The main contributions of the planner are

- The object-oriented workbench architecture to choose and combine different heuristics with different search algorithms and storage structures. The design includes the static analyzer that applies efficient fact-space exploration to distinguish constant from fluent quantities, that clusters facts into groups, and that infers static object symmetries. The static analyzer produces the intermediate format of grounded and simplified planning domain instances.

- Optimal temporal planning enumeration algorithms, based on a precedence relation and PERT scheduling of sequentially generated plans together with a concise analysis of correctness and optimality, as well as the integration of PERT scheduling in MIPS for computing a refined heuristic estimate. This guides the search phase, favoring states with smaller parallel plan length. MIPS instantiates numerical pre- and post-conditions on-the-fly and produces optimized parallel plans.

- The detection of dynamic object symmetries, the integration of different pruning methods such as hash and transposition cuts, as well as different strategies for optimizing objective functions and further implementation tricks that made the system efficient.

The paper analyzes theoretical properties of the contributions, sometimes by slightly abstracting from the actual implementation.

Essentially planning with numerical quantities and durative actions is planning with resources and time. The given framework of mixed propositional and numerical planning problems and the presented intermediate format can be seen as a normal form for temporal and metric planning. The paper presents a novel temporal planning scheme that generates sequential (totally ordered) plans and efficiently schedules them with respect to the set of actions and the imposed causal structure, without falling into known NP-hardness traps for optimized partial-ordering of sequentially generated plans. For smaller problems the complete enumeration approach guarantees optimal solutions. To improve solution quality in approximate enumeration, the (numerical) estimate for the number of operators is replaced by scheduling the relaxed plan in each state. We addressed completeness and optimality of different forms of exploration. A novel study of the time and space complexity of dynamic object symmetry detection is given.

Model checking has always influenced the development of MIPS, e.g in the static analysis to minimize the state description length, in symbolic exploration and plan extraction, in the dependence relation for PERT schedules according to a given partial order, in bit-state hashing for IDA*, in the importance of symmetry detection, an so forth. Moreover, the successes of planning with MIPS can be exported back to model checking, as the development of heuristic search state model checkers and parsing of Promela protocol specifications indicate.

## Acknowledgments

The author would like to thank Derek Long and Maria Fox for helpful discussions concerning this paper and Malte Helmert for his cooperation in the second planning competition. The list of editor's and anonymous reviewers' comments helped a lot to improve the text.

The work is supported by *Deutsche Forschungsgemeinschaft* (DFG) in the projects *Heuristic Search* (Ed 74/3) and *Directed Model Checking* (Ed 74/2).